\documentclass[runningheads]{llncs}

\usepackage{eccv}

\usepackage{eccvabbrv}

\usepackage{graphicx}
\usepackage{booktabs}
\usepackage{wrapfig}
\usepackage{overpic}

\usepackage{multido}
\usepackage{pdfpages}

\usepackage[accsupp]{axessibility}  
\usepackage[dvipsnames]{xcolor}
\usepackage{makecell}

\usepackage{amssymb}
\usepackage{pifont}
\usepackage{lipsum}

\usepackage{tabularx}
\newcolumntype{;}{!{\vrule width 3pt}}
\newcolumntype{:}{!{\vrule width 1.5pt}}

\usepackage{makecell}
\usepackage{colortbl}

\definecolor{firstcolor}{rgb}{1, 0.6, 0.6}
\definecolor{secondcolor}{rgb}{1, 0.8, 0.6}
\definecolor{todocolor}{rgb}{1, 0, 0}
\definecolor{thirdcolor}{rgb}{1,1, 0.6}

\newcommand{\fst}[1]{\cellcolor{firstcolor}#1}
\newcommand{\snd}[1]{\cellcolor{secondcolor}#1}

\usepackage{hyperref}

\usepackage{orcidlink}

\definecolor{higher}{HTML}{9FC4D9} 
\definecolor{lower}{HTML}{f9b0c7}

\begin{document}

\title{LiDAR-Event Stereo Fusion with Hallucinations}

\titlerunning{LiDAR-Event Stereo Fusion}

\author{Luca Bartolomei \orcidlink{0000-0002-5509-437X} \and Matteo Poggi \orcidlink{0000-0002-3337-2236} \and 
Andrea Conti \orcidlink{0000-0002-0197-0178} \and \\ 
Stefano Mattoccia \orcidlink{0000-0002-3681-7704} 
}

\authorrunning{L.~Bartolomei et al.}

\institute{
University of Bologna, Italy \\
\url{https://eventvppstereo.github.io/}}

\maketitle

\begin{figure}
    \centering
    \includegraphics[width=0.95\textwidth]{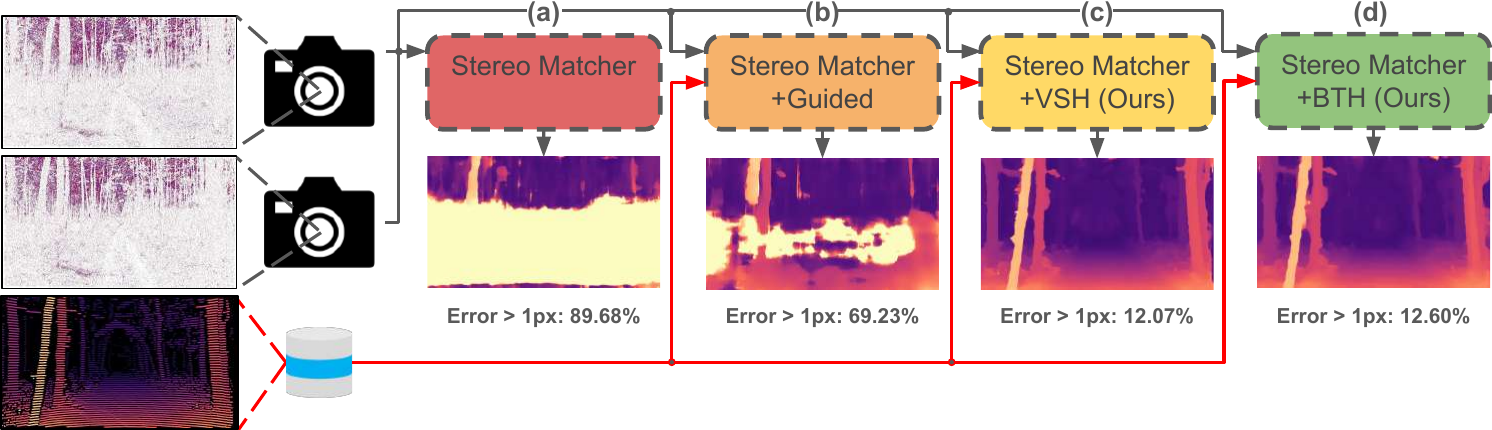}
    \captionof{figure}{\textbf{LiDAR-Event Stereo Fusion with Hallucinations.} In the absence of motion or brightness changes, sparse event streams lead stereo models to catastrophic failures (a). A LiDAR sensor can be used with existing strategies \cite{poggi2019guided} to soften this problem, yet with limited impact (b), whereas our proposals are superior (c,d).
}\label{fig:teaser}
\end{figure}

\begin{abstract}
Event stereo matching is an emerging technique to estimate depth from neuromorphic cameras; however, events are unlikely to trigger in the absence of motion or the presence of large, untextured regions, making the correspondence problem extremely challenging.  
Purposely, we propose integrating a stereo event camera with a fixed-frequency active sensor -- e.g., a LiDAR -- collecting sparse depth measurements, overcoming the aforementioned limitations. 
Such depth hints are used by hallucinating -- i.e., inserting fictitious events -- 
the stacks or raw input streams, compensating for the lack of information in the absence of brightness changes. Our techniques are general, can be adapted to any structured representation to stack events and outperform state-of-the-art fusion methods applied to event-based stereo.

\end{abstract}

\section{Introduction}
\label{sec:intro}

    Depth estimation is a fundamental task with many applications ranging from robotics, 3D reconstruction, and augmented/virtual reality to autonomous vehicles. Accurate, prompt, and high-resolution depth information is crucial for most of these tasks, but obtaining it remains an open challenge. 
    Among the many possibilities, depth-from-stereo is one of the longest-standing approaches to deal with it, with a large literature of deep architecture \cite{poggi2021synergies} proposed in the last decade for processing rectified color images. 
    
    Event cameras \cite{event_survey} (or neuromorphic cameras) are recently emerging as an alternative to overcome the limitations of traditional imaging devices, such as their low dynamic range or the motion blur caused by fast movements. Unlike their traditional counterparts, event cameras do not capture frames at synchronous intervals. Instead, they mimic the dynamic nature of human vision by reporting pixel intensity changes, which can have \textit{positive} or \textit{negative} polarities, as soon as they happen. 
    This peculiarity endows them with unparalleled features -- notably microsecond temporal resolution, and an exceptionally high dynamic range -- making them perfectly suited for applications where fast motion and challenging light conditions are persistent issues (\eg autonomous driving).
    The events streams are often encoded in W$\times$H$\times$C tensors, thus being fully compatible with the CNNs used for classical stereo \cite{nam2022stereo}, capable of estimating dense disparity maps driven by data, despite the sparse nature of events.

    However, as the events trigger only with brightness changes any derived data is \textit{semi-dense} and uninformative, for instance, when facing large untextured regions or in the absence of any motion -- \eg, as in the example in \cref{fig:teaser}. This makes the downstream stereo network struggle to match events across left and right cameras, as shown in \cref{fig:teaser} (a).
    According to the RGB stereo literature, fusing color information with sparse depth measurements from an active sensor \cite{poggi2019guided,cheng2019noise,zhang2022lidar,Bartolomei_2023_ICCV} (\eg, a LiDAR) considerably softens the weaknesses of passive depth sensing, despite the much lower resolution at which depth points are provided. We argue that such a strategy would counter the aforementioned issues even if applied to the event stereo paradigm, yet with a notable nodus caused by the fixed rate at which depth sensors work -- usually, 10Hz for LiDARs -- being in contrast with the asynchronous acquisition rate of event cameras. This would cause to either i) use depth points only when available, harming the accuracy of most fusion strategies known from the classical stereo literature \cite{poggi2019guided,cheng2019noise,zhang2022lidar}, or ii) limiting processing to the LiDAR pace, nullifying one of the greatest strength of event cameras -- \ie, microseconds resolution.
    Nonetheless, this track on event stereo/active sensors fusion has remained unexplored so far. 
    
    In this paper, starting from the RGB literature \cite{poggi2019guided,cheng2019noise,zhang2022lidar,Bartolomei_2023_ICCV}, we embark on a comprehensive investigation into the fusion of event-based stereo with sparse depth hints from active sensors.
    Inspired by \cite{Bartolomei_2023_ICCV}, which projects distinctive color patterns on the images consistently with measured depth, we design a hallucination mechanism to generate fictitious events over time to densify the stream collected by the event cameras. 
    Purposely, we propose two different strategies, respectively consisting of i) creating distinctive patterns directly at the stack level, i.e. a \textbf{Virtual Stack Hallucination} (VSH), just before the deep network processing, or ii) generating raw events directly in the stream, starting from the time instant $t_d$ for which we aim to estimate a disparity map and performing \textbf{Back-in-Time Hallucination} (BTH).
    Both strategies, despite the different constraints -- VSH requires explicit access to the stacked representation, whereas BTH does not -- dramatically improve the accuracy of pure event-based stereo systems, overcoming some of their harshest limitations as shown in Fig. \ref{fig:teaser} (c,d).
    Furthermore, despite depth sensors having a fixed acquisition rate that is in contrast with the asynchronous capture rate of event cameras, VSH and BTH can leverage depth measurements not synchronized with $t_d$ (thus collected at $t_z < t_d$) with marginal drops in accuracy compared to the case of perfectly synchronized depth and event sensors ($t_z = t_d$). This strategy allows for exploiting both VSH and BTH while preserving the microsecond resolution peculiar of event cameras.
    Exhaustive experiments support the following claims:
    
    \begin{itemize}
        \item We prove that LiDAR-stereo fusion frameworks can effectively be adapted to the event stereo domain
        
        \item Our VSH and BTH frameworks are general and work effectively with any structured representation among the eight we surveyed
        
        \item Our strategies outperform existing alternatives inherited from RGB stereo literature on DSEC \cite{Gehrig21ral} and M3ED \cite{Chaney_2023_CVPR} datasets 
        
        \item VSH and BTH can exploit even outdated LiDAR data to increase the event stream distinctiveness and ease matching, preserving the microsecond resolution of event cameras and eliminating the need for synchronous processing dictated by the constant framerate of the depth sensor
        
    \end{itemize}

\section{Related Work}
\label{sec:related_work}

\textbf{Stereo Matching on color images.}
It is a longstanding open problem, with a large body of literature spanning from traditional approaches grounded on handcrafted features and priors \cite{Secaucus_1994_ECCV, veksler2005stereo, yang2008stereo, yang2010constant, liang2011hardware, taniai2014graph, kolmogorov2004energy, hirschmuller2007stereo, boykov2001fast} to contemporary deep learning approaches that brought significant improvements over previous methods, starting with \cite{zbontar2016stereo}.
Nowadays, the most effective solutions have emerged as end-to-end deep stereo networks \cite{poggi2021synergies}, replacing the whole stereo pipeline with a deep neural network architecture through 2D and 3D architectures.
The former, inspired by the U-Net model \cite{ronneberger2015u}, adopts an encoder-decoder design \cite{mayer2016large, Pang_2017_ICCV_Workshops, Liang_2018_CVPR, saikia2019autodispnet, song2018edgestereo, yang2018segstereo, Tonioni_2019_CVPR, yin2019hierarchical, Tankovich_2021_CVPR, Poggi_2024_CVPR_FedStereo}.
In contrast, the latter constructs a feature cost volume from image pair features and estimates the disparity map through 3D convolutions at the cost of substantially higher memory and runtime demands \cite{Kendall_2017_ICCV, chang2018psmnet, khamis2018stereonet, zhang2019ga, cheng2019learning, cheng2020hierarchical, duggal2019deeppruner, yang2019hierarchical, wang2019anytime, guo2019group, Shen_2021_CVPR}. 
A recent trend in this field \cite{lipson2021raft, li2022practical, zhao2023high, zhao2022eai, xu2023iterative, Tosi_2023_CVPR} introduced innovative deep stereo networks that embrace an iterative refinement paradigm or use Vision Transformers \cite{li2021revisiting, guo2022context}. 

\textbf{Stereo Matching with event cameras.}
This topic attracted significant attention due to the unique advantages of event sensors over traditional frame-based cameras.
Similarly to conventional stereo matching, the first approaches focused on developing traditional algorithms by building structured representations, such as voxel grids \cite{schraml2010dynamic}, matched through handcrafted similarity functions \cite{schraml2010dynamic,sulzbachner2011optimized,kogler2011address,zhou2018semi}.
However, pseudo-images lose the high temporal resolution of the stream: to face this problem, \cite{rogister2011asynchronous,carneiro2013event} handle events without an intermediate representation using an event-to-event matching approach, where for each reference event, a set of possible matches is given. 
Camu{\~n}as-Mesa \etal \cite{camunas2014use} add filters to exploit orientation cues and increase matching distinctiveness.
Instead, \cite{piatkowska2013asynchronous} revisited the cooperative network from \cite{marr1976cooperative}.
Neural networks also showed promising results on event stereo matching with models directly processing raw events or using structured representation. 
The former are often inspired by \cite{marr1976cooperative} and typically employ Spiking Neural Networks (SNN) \cite{andreopoulos2018low,dikov2017spiking,osswald2017spiking}. The latter adopts data-driven Convolutional Neural Networks (CNNs) to infer dense depth maps \cite{tulyakov2019learning,uddin2022unsupervised,nam2022stereo}.
A detailed review of different event-based stereo techniques can be found in \cite{gallego2020event}.

\begin{figure}[t]
    \centering
    \includegraphics[trim=0cm 13cm 4cm 0cm, clip,width=0.88\textwidth]{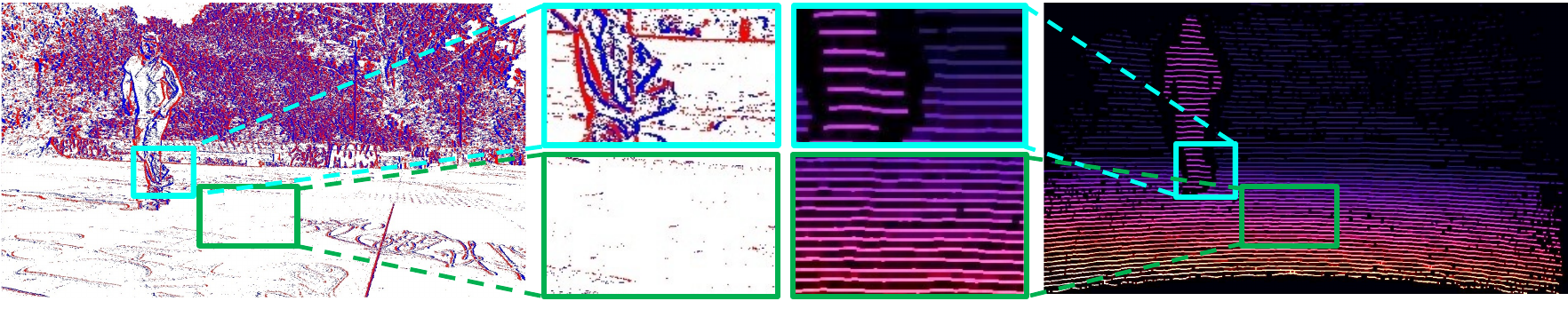}
    \caption{\textbf{Event cameras vs LiDARs -- strengths and weaknesses.} Event cameras provide rich cues at object boundaries where LiDARs cannot (cyan), yet LiDARs can measure depth where the lack of texture makes event cameras uninformative (green). }
    \label{fig:events_vs_lidar}
\end{figure}

\textbf{Sensor fusion for stereo.}
    Recent research has delved into the fusion of color-cameras stereo vision with active sensors, starting with handcrafted algorithms: 
    Badino \etal \cite{Badino2011IntegratingLI} integrated LiDAR data directly into the stereo algorithm using dynamic programming, Gandhi \etal \cite{gandhi2012high} proposed an efficient seed-growing algorithm to fuse time-of-flight (ToF) depth data with stereo pairs, while Marin \etal \cite{MARIN_ECCV_2016} and Poggi \etal \cite{POGGI_IEEE_SENSORS} exploited 
    confidence measures.
    Eventually, contemporary approaches integrated depth from sensors with modern stereo networks, either by concatenating them to images as input \cite{cheng2019noise,park2018high,zhang2020listereo,wang20193d} or by using them to guide the cost optimization process by modulating existing cost volumes \cite{poggi2019guided,huang2021s3,zhang2022lidar,wang20193d}. 
    More recently, Bartolomei \etal \cite{Bartolomei_2023_ICCV} followed a different path with Virtual Pattern Projection (VPP). 
    Although LiDAR sensors and event cameras have been deployed together for some applications \cite{events_completion,brebion2023learning,li2021enhancing,saucedo2023event,song2018calibration,gao2022vector,ta2023l2e}, 
    this paper represents the first attempt at combining LiDAR with an event stereo framework.
    We argue that the two modalities are complementary, as shown in Fig. \ref{fig:events_vs_lidar} -- e.g., the lack of texture and motion makes an event camera uninformative, whereas this does not affect LiDAR systems.

\section{Preliminaries: Event-based Deep Stereo}
\label{sec:prelimnaries}

Event cameras measure brightness changes as an asynchronous stream of events. Accordingly, an event $e_k=(x_k,y_k,p_k,t_k)$ is triggered at time $t_k$ if the intensity sensed by pixel $(x_k,y_k)$ on the W$\times$H sensor grid changes and surpasses a specific contrast threshold. Depending on the sign of this change, it will have polarity $p_k \in \left\{-1,1\right\}$.
Since this unstructured flow is not suitable for standard CNNs -- as those proposed in the classical stereo literature \cite{poggi2021synergies} -- converting it into W$\times$H$\times$C structured representations is necessary if we are interested in obtaining a dense disparity map \cite{Gehrig21ral}. 
Purposely, given a timestamp $t_d$ at which we want to estimate a disparity map, events are sampled backward in time from the stream, either based on a time interval (SBT) or a maximum number of events (SBN),  and \textit{stacked} according to various strategies -- among them: 

\textbf{Histogram} \cite{maqueda2018event}. Events of the two polarities are counted into per-pixel histograms, yielding a W$\times$H$\times$2 stack.

\textbf{Voxel grid} \cite{zhu2019unsupervised}. The timelapse from which events are sampled is split into $B$ uniform bins: polarities are accumulated in each bin of a W$\times$H$\times$B stack.

\textbf{Mixed-Density Event Stack (MDES)} \cite{nam2022stereo}. Similar to the voxel grid strategy, the timelapse is split into bins covering $1, \frac{1}{2}, \frac{1}{4}, ..., \frac{1}{2^{N-2}}, \frac{1}{2^{N-1}}$ of the total interval. The latest event in each bin is kept, yielding a W$\times$H$\times$N binary stack.

\textbf{Concentrated stack} \cite{nam2022stereo}. A shallow CNN is trained to process a pre-computed stack (\eg, an MDES) and aggregate it to a W$\times$H$\times$1 data structure.

\textbf{Time-Ordered Recent Event (TORE)} \cite{baldwin2022time}. It stores
event timestamps into Q per-pixel queues for each polarity, yielding a W$\times$H$\times$2Q stack.

\textbf{Time Surface} \cite{lagorce2016hots}. A surface is derived from the timestamp distributions of the two polarities. S values are sampled for each, yielding a W$\times$H$\times$2S stack.

\textbf{ERGO-12} \cite{Zubic_2023_ICCV}. An optimized representation of 12 channels, each built according to different strategies from the previous. It yields a W$\times$H$\times$12 stack.

\textbf{Tencode} \cite{Huang_2023_WACV}. A color image representation in which R and B channels encode positive and negative polarities, with G encoding the timestamp relative to the total timelapse. It produces an RGB image, \ie a W$\times$H$\times$3 stack.

We can broadly classify stereo frameworks using these representations into three categories: i) \textit{white boxes}, for which we have full access to the implementation of both the stereo backbone and the stacked event construction; ii) \textit{gray boxes}, in case we do not have access to the stereo backbone; iii) \textit{black boxes}, when the stacked event representation is not accessible neither.

\section{Proposed Method}
\label{sec:method}

According to the sensor fusion literature for conventional cameras, the main strategies for combining stereo images with sparse depth measurements from active sensors consist of i) concatenating the two modalities and processing them as joint inputs with a stereo network \cite{cheng2019noise,park2018high,zhang2020listereo,wang20193d}, ii) modulating the internal cost volume computed by the backbone itself \cite{poggi2019guided,huang2021s3,zhang2022lidar,wang20193d} or, more recently, iii) projecting distinctive patterns on images according to depth hints \cite{Bartolomei_2023_ICCV}.

\begin{figure}[t]
    \centering
    \begin{tabular}{cc:ccc}
     \includegraphics[width=0.3\textwidth]{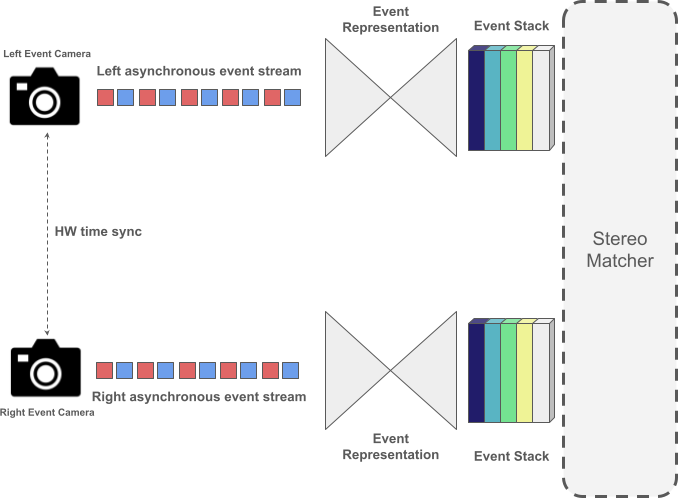} & \quad &
     \includegraphics[width=0.3\textwidth]{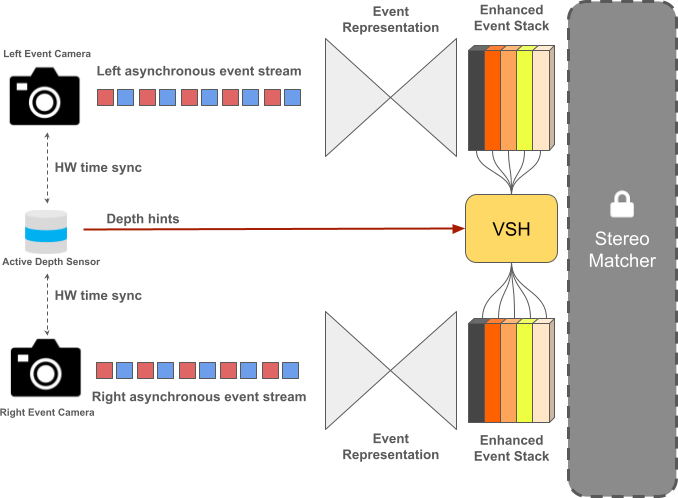} & \quad &
     \includegraphics[width=0.3\textwidth]{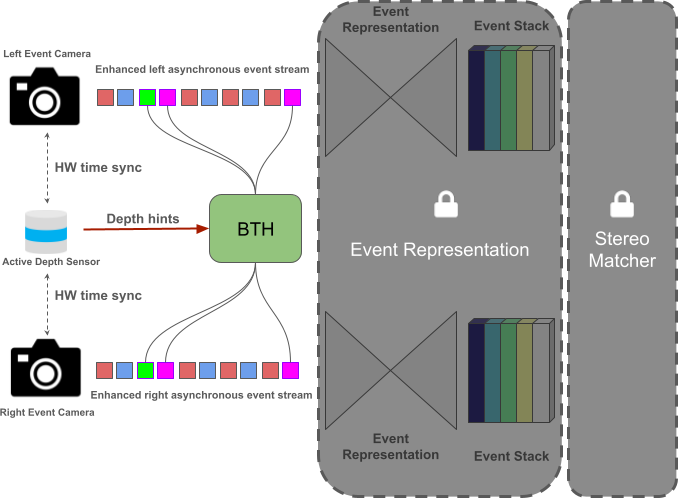}
     \\
     \tiny\textbf{(a)} & \multicolumn{1}{c}{} & \tiny\textbf{(b)} & & \tiny\textbf{(c)} \\
    \end{tabular}
    \caption{\textbf{Overview of a generic event-based stereo network and our hallucination strategies.} State-of-the-art event-stereo frameworks (a) pre-process raw events to obtain event stacks fed to a deep network. In case the stacks are accessible, we define the model as a \textit{gray box}, otherwise as a \textit{black box}.
    In the former case (b), we can hallucinate patterns directly on it (VSH). When dealing with a black box (c), we can hallucinate raw events that will be processed to obtain the stacks (BTH).}
    \label{fig:vsh_bth}
\end{figure}

We follow the latter path, 
since it is more effective and flexible than the alternatives -- which can indeed be applied to \textit{white box} frameworks only. 
For this purpose, we design two alternative strategies suited even for gray and black box frameworks, respectively, as depicted in Fig. \ref{fig:vsh_bth}.

\subsection{\underline{V}irtual \underline{S}tack \underline{H}allucination -- VSH} 

Given left and right stacks $\mathcal{S}_L,\mathcal{S}_R$ of size W$\times$H$\times$C and a set $Z$ of depth measurements $z(x,y)$ by a sensor, we perform a Virtual Stack Hallucination (VSH), by augmenting each channel $c\in\text{C}$, to increase the distinctiveness of local patterns and thus ease matching. This is carried out by injecting the same virtual stack $\mathcal{A}(x,y,x',c)$ into $\mathcal{S}_L,\mathcal{S}_R$ respectively at coordinates $(x,y)$ and $(x',y)$.

\begin{equation}
    \begin{split}
        \mathcal{S}_L(x,y,c) \leftarrow \mathcal{A}(x,y,x',c)\\
        \mathcal{S}_R(x',y,c) \leftarrow \mathcal{A}(x,y,x',c)
    \end{split}
    \label{eq:esvpp_principle}
\end{equation}
with $x'$ obtained as $x-d(x,y)$, with disparity $d(x,y)$ triangulated back from depth $z(x,y)$ as $\frac{bf}{z(x,y)}$, according to the baseline and focal lengths $b,f$ of the stereo system.
We deploy a generalized version of the random pattern operator $\mathcal{A}$ proposed in \cite{Bartolomei_2023_ICCV}, agnostic to the stacked representation:

\begin{equation}
    \mathcal{A}(x,y,x',c) \sim \mathcal{U}(\mathcal{S}^-, \mathcal{S}^+)
    \label{eq:esvpp_random}
\end{equation}
with $\mathcal{S}^-$ and $\mathcal{S}^+$ the minimum and maximum values appearing across stacks $\mathcal{S}_L,\mathcal{S}_R$ and $\mathcal{U}$ a uniform random distribution. Following \cite{Bartolomei_2023_ICCV}, the pattern can either cover a single pixel or a local window.
This strategy alone is sufficient already to ensure distinctiveness and to dramatically ease matching across stacks, even more than with color images \cite{Bartolomei_2023_ICCV}, since acting on semi-dense structures -- \ie, stacks are uninformative in the absence of events. It also ensures a straightforward application of the same principles used on RGB images, \eg, to combine the original content (color) with the virtual projection (pattern) employing alpha blending \cite{Bartolomei_2023_ICCV}.
Nevertheless, we argue that acting at this level i) requires direct access to the stacks, i.e., a gray-box deep event-stereo network, and ii) might be sub-optimal as stacks encode only part of the information from streams.

\subsection{\underline{B}ack-in-\underline{T}ime \underline{H}allucination -- BTH}

A higher distinctiveness to ease correspondence can be induced by hallucinating patterns directly in the continuous events domain.
Specifically, we act in the so-called \textit{event history}: given a timestamp $t_d$ at which we want to estimate disparity, raw events are sampled from the left and right streams starting from $t_d$ and going backward, according to either SBN or SBT stacking approaches, to obtain a pair of event histories $\mathcal{E}_L = \left\{ e^L_k \right\}^{N}_{k=1}$ and $\mathcal{E}_R = \left\{ e^R_k \right\}^{M}_{k=1}$,
where $e^L_k,e^R_k$ are the $k$-th left and right events.
Events in the history are sorted according to their timestamp -- \ie, inequality $t_k \leq t_{k+1}$ holds for every two adjacent $e_{k},e_{k+1}$.

\begin{figure*}[t]
    \centering
    \includegraphics[trim=0.5cm 12cm 4cm 4cm,clip,width=0.9\textwidth]{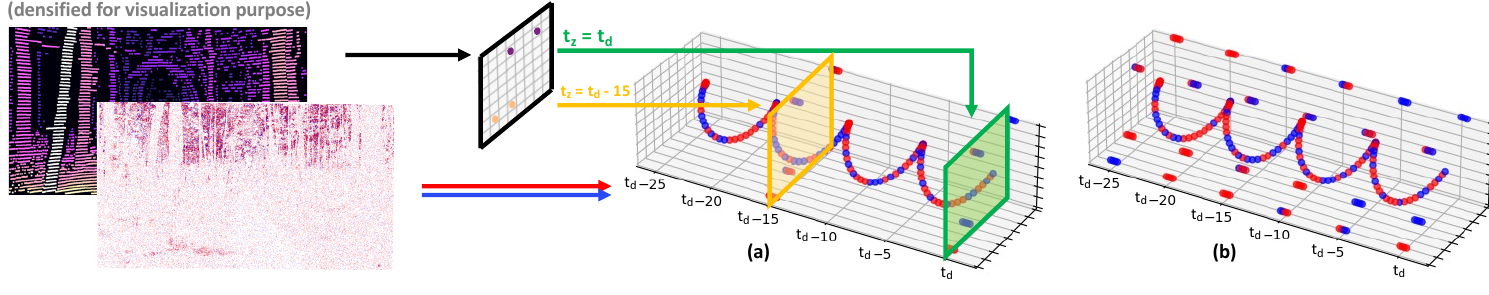}
    \caption{\textbf{Overview of Back-in-Time Hallucination (BTH).} To estimate disparity at $t_d$, if LiDAR data is available -- e.g., at timestamp $t_z=t_d$ (green) or $t_z=t_d-15$ (yellow) -- we can na\"ively inject events of random polarities at the same timestamp $t_z$ (a). 
    More advanced injection strategies can be used -- \eg by hallucinating multiple events, starting from $t_d$, back-in-time at regular intervals (b). }
    \label{fig:bth}
\end{figure*}

At this point, we intervene to hallucinate novel events:
given a depth measurement $z(\hat{x},\hat{y})$, triangulated back into disparity $d(\hat{x},\hat{y})$, we inject a pair of fictitious events $\hat{e}^L=(\hat{x},\hat{y},\hat{p},\hat{t})$ and $\hat{e}^R=(\hat{x}',\hat{y},\hat{p},\hat{t})$ respectively inside $\mathcal{E}_L$ and $\mathcal{E}_R$, producing $\hat{\mathcal{E}}_L=\left\{e^L_1,\dots,\hat{e}^L,\dots,e^L_N\right\}$ and $\hat{\mathcal{E}}_R=\left\{e^R_1,\dots,\hat{e}^R,\dots,e^R_M\right\}$. 
By construction, $\hat{e}^L$ and $\hat{e}^R$ adhere to i) the time ordering constraint, ii) the geometry constraint $\hat{x}'=\hat{x}-d(\hat{x},\hat{y})$ and iii) a similarity constraint -- \ie, $\hat{p},\hat{t}$ are the same for $\hat{e}^L$ and $\hat{e}^R$.
Fictitious polarity $\hat{p}$ and fictitious timestamp $\hat{t}$ are two degrees of freedom useful to ensure distinctiveness along the epipolar line and ease matching, according to which we can implement different strategies summarized in Fig. \ref{fig:bth}, and detailed in the remainder.

\textbf{Single-timestamp injection.}
The simplest way to increase distinctiveness is to insert synchronized events at a fixed timestamp.
Accordingly, for each depth measurement $d(\hat{x},\hat{y})$, a total of $K_{\hat{x},\hat{y}}$ pairs of fictitious events are inserted in $\mathcal{E}_L,\mathcal{E}_R$, having polarity $\hat{p}_k$ randomly chosen from the discrete set $\left\{-1,1\right\}$. Timestamp $\hat{t}$ is fixed and can be, for instance, $t_z$ at which the sensor infers depth, that can coincide with timestamp $t_d$ at which we want to estimate disparity -- \eg, $t_z=t_d=0$ in the case depicted in Fig. \ref{fig:bth} (a). Inspired by \cite{Bartolomei_2023_ICCV}, events might be optionally hallucinated in patches rather than single pixels.
However, as depth sensors usually work at a fixed acquisition frequency -- \eg, 10Hz for LiDARs -- sparse points might be unavailable at any specific timestamp. Nonetheless, since $\mathcal{E}_L,\mathcal{E}_R$ encode a time interval, we can hallucinate events even if derived from depth scans performed \textit{in the past} -- \eg, at $t_z<t_d$, 
-- by placing them in the proper position inside $\mathcal{E}_L,\mathcal{E}_R$.

\textbf{Repeated injection.}
The previous strategy does not exploit one of the main advantages of events over color images, \ie the temporal dimension, at its best. Purposely, we design a more advanced hallucination strategy based on \textit{repeated} na\"ive injections performed along the time interval sampled by $\mathcal{E}_L, \mathcal{E}_R$. 
As long as we are interested in recovering depth at $t_d$ only, we can hallucinate as many events as we want in the time interval \textit{before} $t$ -- \ie, for $t_z=t_d=0$, over the entire interval as shown in Fig. \ref{fig:bth} (b) -- consistent with the depth measurements at $t_d$ itself, which will increase the distinctiveness in the event histories and will ease the match by hinting the correct disparity. 
Inspired by the stacked representations introduced in Sec. \ref{sec:prelimnaries}, we can design a strategy for injecting multiple events along the stream. 
Accordingly, we define the \textit{conservative} time range $\left[t^-,t^+\right]$ of the events histories $\mathcal{E}_L, \mathcal{E}_R$,
with $t^-=\min\left\{t^L_0,t^R_0\right\}$ and $t^+=\max\left\{t^L_N,t^R_M\right\}$
and divide it into $B$ equal temporal bins.
Then, inspired by MDES \cite{nam2022stereo}, we run $B$ single-timestamp injections at 
$\hat{t}_b=\frac{2^b-1}{2^b}(t^+-t^-)+t^-$, with $b \in \left\{1,\dots,B\right\}$. 
Additionally, each depth measurement is used only once -- \ie, the number of fictitious events $K_{b,\hat{x},\hat{y}}$ in the $b$-th injection is set as $K_{b,\hat{x},\hat{y}} \leftarrow K_{\hat{x},\hat{y}}\delta(b,D_{\hat{x},\hat{y}})$ where $\delta(\cdot,\cdot)$ is the Kronecker delta and $D_{\hat{x},\hat{y}}\leftarrow\text{round}(X^\mathcal{U}(B-1)+1)$ is a random slot assignment.
We will show in our experiment how this simple strategy can improve the results of BTH, in particular increasing its robustness against misaligned LiDAR data -- i.e., measurements retrieved at a timestamp $t_z < t_d$.

\section{Experiments}
\label{sec:experiments}

\subsection{Implementation and Experimental Settings}

We implement VSH and BTH in Python, using the Numba package. 

\textbf{General framework.} We build our code base starting from SE-CFF \cite{nam2022stereo} -- state-of-the-art for event-based stereo -- assuming the same stereo backbone as in their experiments, \ie derived from AANet \cite{xu2020aanet}, and run SBN to generate the event history to be stacked. While we select a single architecture, we implement a variety of stacked representations: purposely, we implement a single instance of the stereo backbone for any stacked representations introduced in \cref{sec:prelimnaries}, taking the opportunity to evaluate their performance with the event stereo task. For Concentration representation, we use MDES as the prior stacking function following \cite{nam2022stereo} and avoid
considering future events during training. Furthermore, in this case, VSH is applied before the concentration network since it would interfere with gradient back-propagation during training -- while this cannot occur with BTH.
From \cite{Bartolomei_2023_ICCV}, we adapt occlusion handling and hallucination on uniform/not uniform patches. We also implement alpha-blending, for VSH only -- as it loses its purpose when acting on the raw streams. For all our methods, we inherit the same hyper-parameters from \cite{Bartolomei_2023_ICCV}; yet, we discard occluded points as the occlusion handling strategy for BTH since an equivalent strategy to deal with sparse event histories is not trivial.
For VSH on Voxel Grids, we use the 5-th and 95-th percentile to calculate $\mathcal{S}^-$ and $\mathcal{S}^+$ due to the frequent presence of extreme values in the stack. For BTH, we perform 12 injections (\ie, $B=12$).

\textbf{Existing fusion methodologies.} We compare our proposal with existing methods from the RGB stereo literature, consisting of i) modulating the cost volume built by the backbone -- Guided Stereo Matching \cite{poggi2019guided}, ii) concatenating the sparse depth values to the inputs to the stereo network -- \eg, as done by LidarStereoNet \cite{cheng2019noise}, iii) a combination of both the previous strategies -- in analogy to CCVNorm \cite{wang20193d}. 
Any strategy is adapted to the same common stereo backbone \cite{nam2022stereo} 
(see \textbf{supplementary material}). Running BTH and VSH adds respectively 10ms and 2-15ms (depending on representations) on the CPU.

\textbf{Training protocol.} Any model we train -- either the original event stereo backbones or those implementing fusion strategies -- runs for 25 epochs with a batch size of 4 and a maximum disparity set to 192.
We use Adam \cite{kingma2014adam} with beta (0.9, 0.999) 
and weight decay set to $10^{-4}$.
The learning rate starts at $5\cdot10^{-4}$ and decays with cosine annealing. 
We apply random crops and vertical flips to augment data during training.

\begin{figure}[t]
    \centering
    \renewcommand{\tabcolsep}{1pt}
    \begin{tabular}{cccc}
        \multicolumn{2}{c}{\tiny DSEC \cite{Gehrig21ral}} & \multicolumn{2}{c}{\tiny M3ED \cite{Chaney_2023_CVPR}} \\
        \includegraphics[trim=0cm 2cm 0cm 0cm, clip, height=0.15\linewidth]{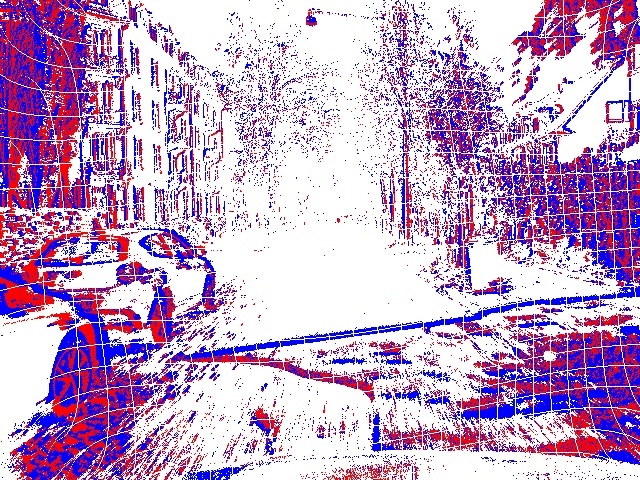} 
        & 
        \includegraphics[trim=0cm 2cm 0cm 0cm, clip, height=0.15\linewidth]{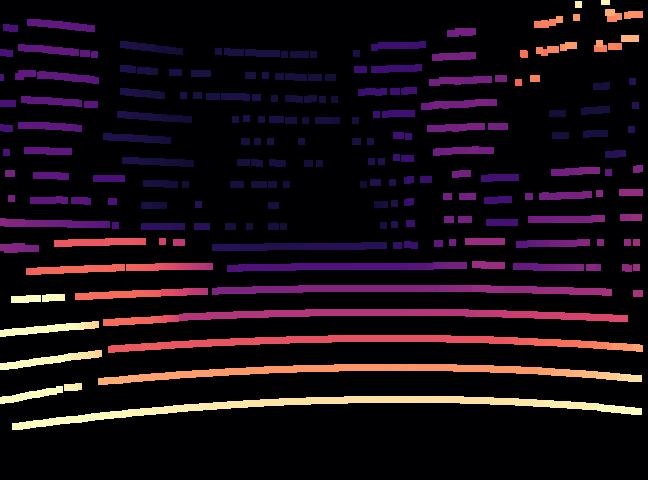} &
        \includegraphics[height=0.15\linewidth]{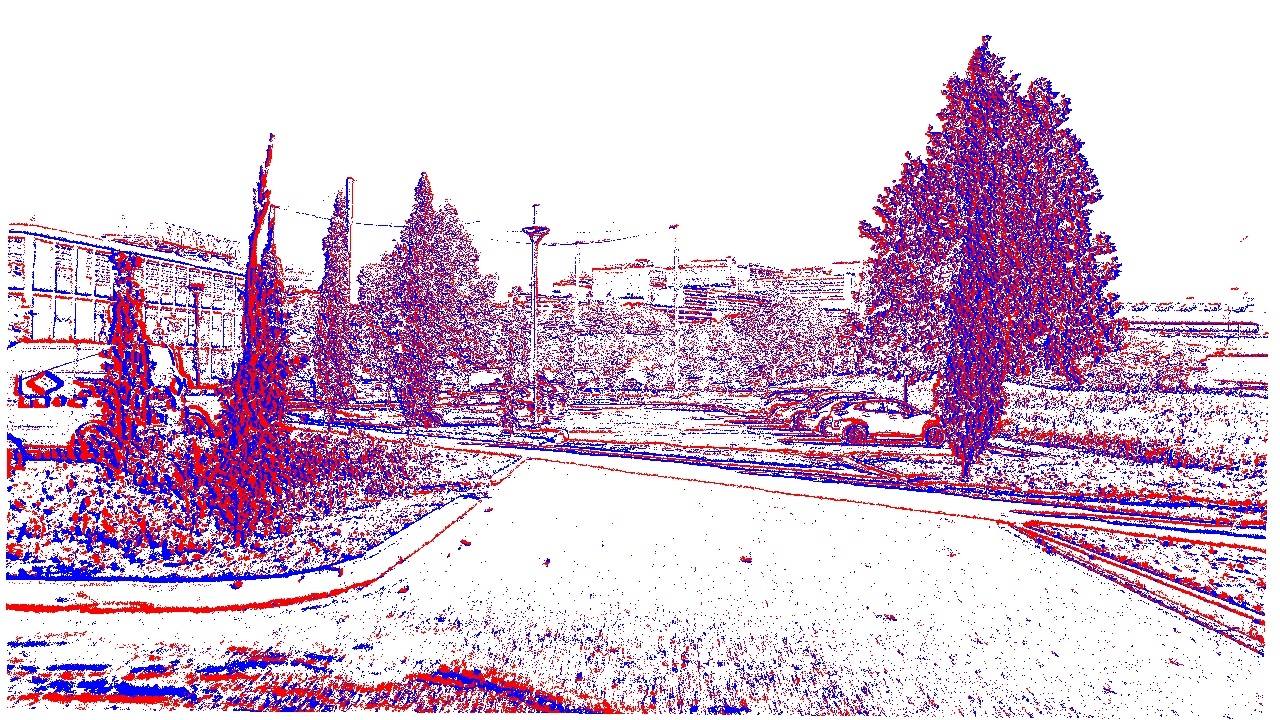} 
        & 
        \includegraphics[height=0.15\linewidth]{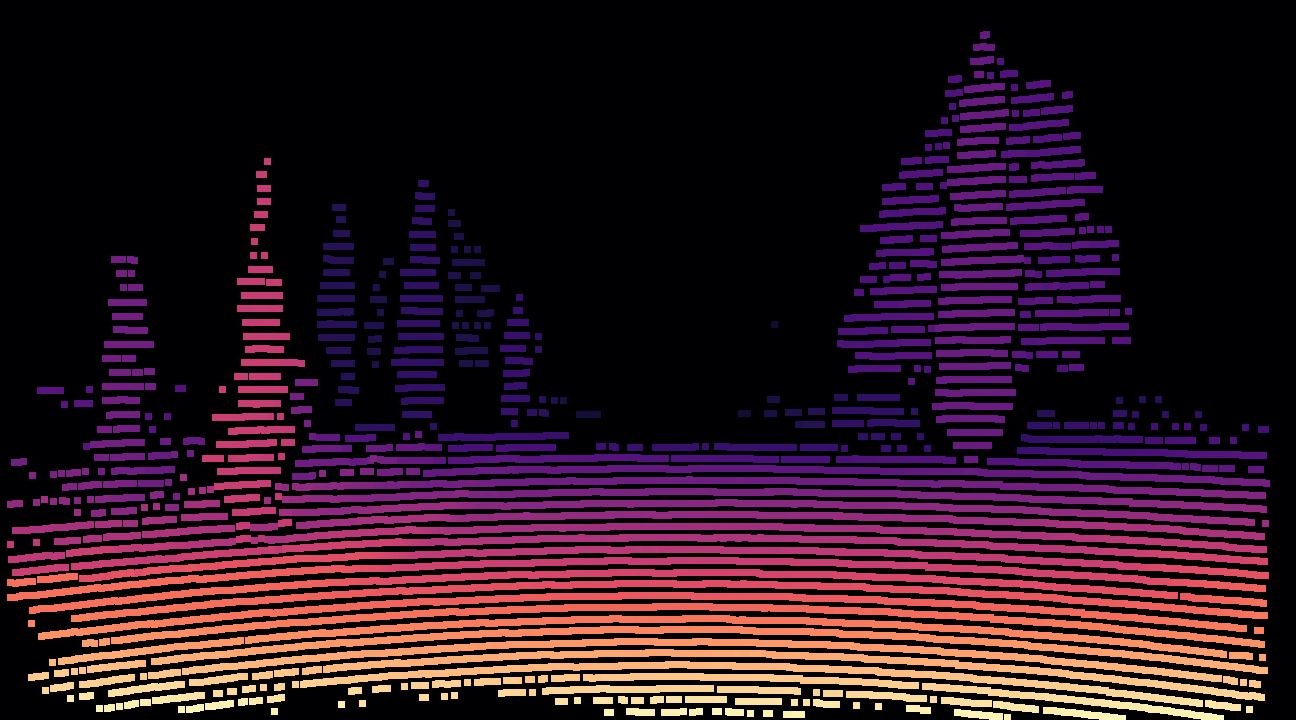} \\
    \end{tabular}
    \caption{\textbf{Qualitative comparison -- DSEC vs M3ED.} DSEC features $640\times480$ event cameras and a 16-line LiDAR, M3ED has $1280\times720$ event cameras and a 64-line LiDAR. LiDAR scans have been dilated with a $7\times7$ kernel to ease visualization. }
    \label{fig:dsec_vs_m3ed}
\end{figure}

\subsection{Evaluation Datasets \& Protocol}
\label{sec:eval_dataset_protocol}

We introduce datasets and metrics used in our experiments. 

\textbf{DSEC \cite{Gehrig21ral}.} An outdoor event stereo dataset, captured using wide-baseline (50 cm) stereo event cameras at $640\times480$ resolution. Ground-truth disparity is obtained by accumulating 16-line LiDAR scans, for a total of 26\,384 maps organized into 41 sequences. 
We split them into train/test sets following \cite{nam2022stereo}. From the training set, we retain a further \textit{search} split for hyper-parameters tuning and ablation experiments. 
Sparse LiDAR measurements are obtained by aligning the raw scans with the ground-truth -- both provided by the authors -- by running a LiDAR inertial odometry pipeline followed by ICP registration (see the \textbf{supplementary material} for details).

\textbf{M3ED \cite{Chaney_2023_CVPR}.} This dataset provides 57 indoor/outdoor scenes collected with a compact multi-sensor block mounted on three different vehicles -- \ie, a car, a UAV, and a quadruped robot.
A 64-line LiDAR generates semi-dense ground-truth depth, while the event stereo camera has a shorter baseline (12 cm) and a higher resolution ($1280\times720$). 
We use 5 sequences from this dataset for evaluation purposes only -- some of which contain several frames acquired with the cameras being static -- to evaluate the generalization capacity of the models both to different domains and the density of the LiDAR sensor. 
Similarly to DSEC, we derived sparse LiDAR depth maps from the raw scans. Thanks to the SDK made available by the authors, we could derive LiDAR measurements aligned to any desired temporal offset according to linear interpolation of the ground-truth poses (see the \textbf{supplementary material} for details). This allows us to run dedicated experiments to assess the effect of time-misaligned depth measurements.
Fig. \ref{fig:dsec_vs_m3ed} shows a qualitative comparison between the two datasets.

\textbf{Evaluation Metrics.} We compute the percentage of pixels with an error greater than 1 or 2 pixels (1PE, 2PE), and the mean absolute error (MAE). We highlight the \colorbox{firstcolor}{\textbf{best}} and \colorbox{secondcolor}{\textbf{second best}} methods per row on each metric. 

\begin{figure*}[t]
    \centering
    \includegraphics[width=0.48\textwidth]{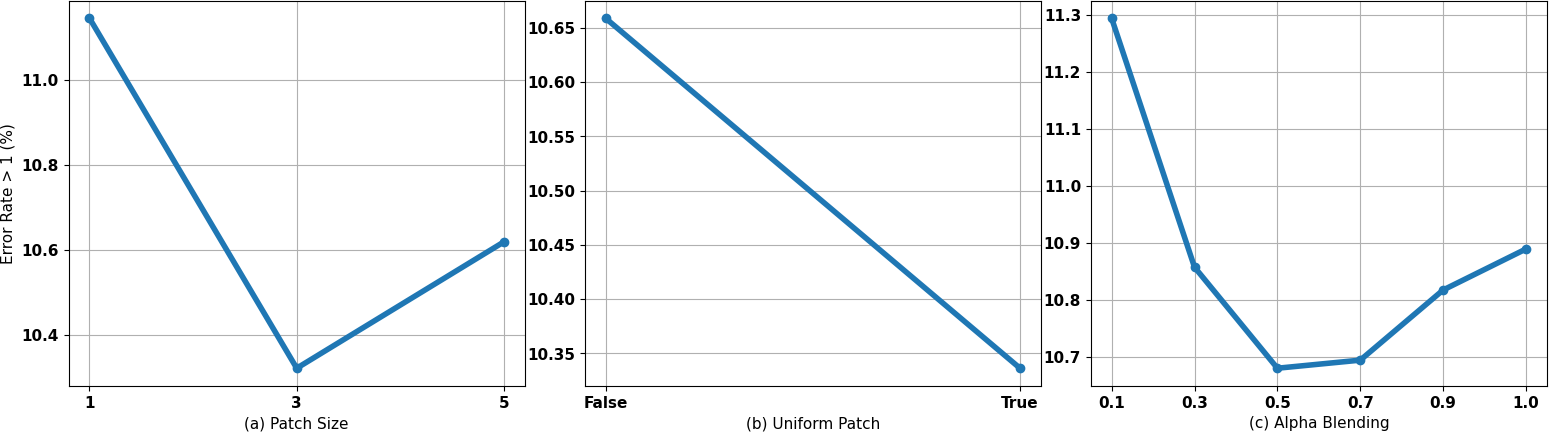}
    \includegraphics[width=0.8\textwidth]{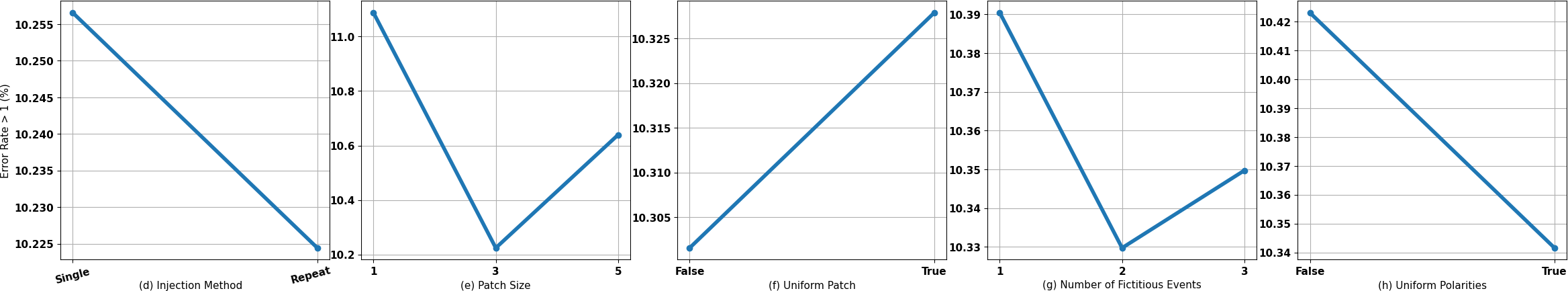}
    \caption{\textbf{Hyperparameters search.} Results on DSEC search split. On top, we study the impact of (a) patch size, (b) uniform patches, and (c) alpha blending on VSH. At the bottom, we consider 
    (d) single vs repeated injection, (e) patch size, (f) uniform patches, (g) number of fictitious events, and (h) uniform polarities on BTH.}
    \label{fig:bth_method_search}
\end{figure*}

\subsection{Ablation Study}
\label{sec:ablation_study}

We ran hyper-parameters search and ablation experiments for VSH and BTH on the DSEC search split, reporting the 1PE error.
We conducted these experiments using any representation listed in \cref{sec:prelimnaries} -- except for Concentration \cite{nam2022stereo}, which starts from pre-computed MDES stacks -- and report the average results.

\textbf{VSH.} 
\cref{fig:bth_method_search} (top) shows the impact of different hyper-parameters on VSH strategy. 
In (a), we can observe how VSH is improved by using $3\times3$ patches, while $5\times5$ cannot yield further benefits. Consequently, we select it as the default configuration from now on. In (b), we show that uniform patterns are more effective than random ones, and in (c) alpha equal to 0.5 works the best.

\textbf{BTH.} \cref{fig:bth_method_search} (bottom) focuses on our second strategy.
In (d) we show how repeated injection can improve the results; thus, we select it as the default configuration from now on. In the remainder, we will better appreciate how this setting is much more robust when dealing with misaligned LiDAR data.
Next, (e) outlines how hallucinating events with $3\times3$ patches lead to the best results. Applying a uniform patch of events following (f) yields, again, better results.
In (g), we tested different numbers $K_{\hat{x},\hat{y}}$ of injected fictitious events. Injecting more than one event is beneficial, yet saturating with two. Finally, (h) shows that using uniform polarities yields lower errors.

\begin{table*}[t]
    \centering
    \caption{\textbf{Results on DSEC \cite{Gehrig21ral} -- pre-trained.} We test the different stacked representations (rows) with several fusion strategies applied to pre-trained stereo backbones.}
    \renewcommand{\tabcolsep}{7pt}
    \scalebox{0.6}{
    \begin{tabular}{c}

    \hspace{-0.4cm}\begin{tabular}{crrr:rrr:rrr:rrr:rrr}
        \Xhline{3pt}
        & \multicolumn{3}{l;}{Stacked} & \multicolumn{3}{c:}{Baseline} & \multicolumn{3}{c:}{Guided \cite{poggi2019guided}} & \multicolumn{3}{c:}{VSH (ours)} & \multicolumn{3}{c}{BTH (ours)} \\
        & \multicolumn{3}{l;}{representation} & 1PE & 2PE & MAE & 1PE & 2PE & MAE & 1PE & 2PE & MAE & 1PE & 2PE & MAE \\
        \Xhline{3pt}
        (A) & \multicolumn{3}{l;}{Histogram \cite{maqueda2018event}} & 16.21 & 4.73 & 0.74 & 16.07 & 4.68 & 0.73 & \snd 13.71 & \snd 4.20 & \snd 0.69 & \fst 13.32 & \fst 3.92 & \fst 0.66 \\
        (B) & \multicolumn{3}{l;}{MDES \cite{nam2022stereo}} & 15.32 & 4.40 & 0.70 & 15.13 & 4.34 & 0.70 & \snd 12.94 & \snd 3.52 & \snd 0.63 & \fst 12.61 & \fst 3.50 & \fst 0.62 \\
        (C) & \multicolumn{3}{l;}{Concentration \cite{nam2022stereo}} & 15.97 & 4.33 & 0.70 & 15.79 & 4.27 & 0.70 &\fst  13.70 & \fst 3.60 & \fst 0.65 & \snd 14.66 & \snd 3.77 & \snd 0.66 \\
        (D) & \multicolumn{3}{l;}{Voxelgrid \cite{zhu2019unsupervised}} & 16.49 & 4.56 & 0.72 & 16.29 & 4.50 & 0.71 & \snd 13.12 & \snd 3.69 & \snd 0.65 & \fst 12.44 & \fst 3.60 & \fst 0.62 \\
        (E) & \multicolumn{3}{l;}{TORE \cite{baldwin2022time}} & 15.91 & 4.57 & 0.71 & 15.72 & 4.50 & 0.71 & \snd 12.53 & \fst 3.65 & \snd 0.63 & \fst 12.27 & \snd 3.68 & \fst 0.62 \\
        (F) & \multicolumn{3}{l;}{Time Surface \cite{lagorce2016hots}} & 15.33 & 4.29 & 0.70 & 15.18 & 4.24 & 0.69 & \fst 12.16 & \fst 3.38 & \fst 0.62 & \snd 12.28 & \snd 3.45 & \fst 0.62 \\
        (G) & \multicolumn{3}{l;}{ERGO-12 \cite{Zubic_2023_ICCV}} & 15.02 & 4.20 & 0.68 & 14.87 & 4.14 & 0.68 & \snd 12.02 & \fst 3.40 & \fst 0.61 & \fst 11.98 & \snd 3.42 & \fst 0.61 \\
        (H) & \multicolumn{3}{l;}{Tencode \cite{Huang_2023_WACV}} & 14.46 & 4.17 & 0.68 & 14.29 & 4.11 & 0.67 & \snd 12.12 & \fst 3.37 & \fst 0.61 & \fst 11.86 & \snd 3.45 & \fst 0.61 \\
        \Xhline{3pt}
        & \multicolumn{3}{l;}{Avg. Rank.} & \multicolumn{3}{c:}{-} & 3.00 & 3.00 & 3.00 & \snd 1.75 & \fst 1.38 & \snd 1.50 & \fst 1.25 & \snd 1.63 & \fst 1.13 \\
        \Xhline{3pt}
    \end{tabular}    

    \end{tabular}
    }
    \label{tab:dsec_notrain}
\end{table*}

\begin{table*}[t]
    \centering
    \caption{\textbf{Results on DSEC \cite{Gehrig21ral} -- retrained.} We test different stacked representations (rows) with several fusion strategies applied during training.}
    \renewcommand{\tabcolsep}{7pt}
    \scalebox{0.6}{
    \begin{tabular}{c}

    \hspace{-0.4cm}\begin{tabular}{crrr:rrr:rrr:rrr:rrr}
        \Xhline{3pt}
        & \multicolumn{3}{c:}{Concat \cite{cheng2019noise}} & \multicolumn{3}{c:}{Guided+Concat \cite{wang20193d}} & \multicolumn{3}{c:}{Guided \cite{poggi2019guided}} & \multicolumn{3}{c:}{VSH (ours)} & \multicolumn{3}{c}{BTH (ours)} \\ 
        & 1PE & 2PE & \multicolumn{1}{r:}{MAE}  & 1PE & 2PE & \multicolumn{1}{r:}{MAE} & 1PE & 2PE & MAE & 1PE & 2PE & MAE & 1PE & 2PE & MAE \\
        \Xhline{3pt}
        (A) & 12.57 & \snd 3.37 & 0.62 & 12.81 & 3.41 & 0.63 & 15.57 & 4.58 & 0.72 & \fst 9.90 & \fst 3.26 & \fst 0.53 & \snd 10.91 & 3.41 & \snd 0.59 \\
        (B) & 12.37 & 3.17 & 0.61 & 12.40 & 3.25 & 0.61 & 14.66 & 4.36 & 0.70 & \fst 9.31 & \fst 3.01 & \fst 0.51 & \snd 9.62 & \fst 3.01 & \snd 0.54 \\
        (C) & 12.38 & 3.41 & 0.63 & 12.74 & 3.44 & 0.66 & 15.15 & 4.45 & 0.71 & \snd 9.70 & \snd 3.04 & \fst 0.53 & \fst 9.66 & \fst 2.98 & \snd 0.55 \\
        (D) & 12.23 & 3.18 & 0.60 & 11.90 & \snd 3.10 & 0.60 & 14.52 & 4.21 & 0.68 & \snd 10.16 & 3.20 & \snd 0.56 & \fst 9.68 & \fst 2.90 & \fst 0.54 \\
        (E) & 12.99 & 3.33 & 0.62 & 12.62 & 3.25 & 0.61 & 16.00 & 4.56 & 0.73 & \snd 9.91 & \snd 3.05 & \fst 0.53 & \fst 9.83 & \fst 2.98 & \snd 0.54\\
        (F) & 12.18 & 3.09 & 0.61 & 12.47 & 3.17 & 0.61 & 14.40 & 4.21 & 0.68 & \fst 9.47 & \fst 2.90 & \fst 0.52 & \snd 9.58 & \snd 2.92 & \snd 0.54 \\
        (G) & 12.43 & 3.14 & 0.61 & 12.82 & 3.19 & 0.62 & 13.85 & 3.97 & 0.66 & \fst 9.25 & \snd 2.88 & \fst 0.50 & \snd 9.37 & \fst 2.87 & \snd 0.54\\
        (H) & 11.95 & 3.08 & 0.60 & 11.75 & 3.10 & 0.60 & 14.72 & 4.21 & 0.69 & \fst 9.39 & \snd 3.00 & \fst 0.52 & \snd 9.59 & \fst 2.97 & \snd 0.55\\
        \Xhline{3pt}
        
        & 3.38 & 3.00 & 3.13 & 3.63 & 3.50 & 3.38 & 5.00 & 5.00 & 5.00 & \fst 1.38 & \snd 1.88 & \fst 1.13 & \snd 1.63 & \fst 1.38 & \snd 1.88 \\
        \Xhline{3pt}
    \end{tabular}    

    \end{tabular}
    }
    \label{tab:dsec_train}
\end{table*}

\subsection{Experiments on DSEC}\label{sec:dsec}

We now report experiments on the DSEC testing split, either when applying fusion strategies to pre-trained stereo models without retraining them or when training the networks from scratch to exploit LiDAR data.

\textbf{Pre-trained models.} \cref{tab:dsec_notrain} reports, on each row, the results yielded by using a specific stacked representation. In the columns, we report the different fusion strategies involved in our experiments, starting with the baseline -- \ie, a stereo backbone processing events only.
In the last row, we report the average ranking -- for the three metrics -- achieved by any fusion strategy over the eight representations.
Starting from baseline models, we can notice how the different representations have an impact on the accuracy of the stereo backbone, with those modeling complex behaviors -- \eg, Time Surface \cite{lagorce2016hots} or ERGO-12 \cite{Zubic_2023_ICCV} -- yielding up to 2\% lower 1PE than simpler ones such as Histogram \cite{maqueda2018event}.
The Guided framework \cite{poggi2019guided} can improve the results only moderately: this is caused by the very sparse measurements retrieved from the 16-line LiDAR sensor used in DSEC, as well as by the limited effect of the cost volume modulation in regions where events are not available for matching. Nonetheless, VSH and BTH consistently outperform Guided, always improving the baseline by 2-3\% points on 1PE. In general, BTH achieves the best 1PE and MAE metrics in most cases;
this strategy is the best when re-training the stereo backbone is not feasible. 

\textbf{Training from scratch.} \cref{tab:dsec_train} reports the results obtained by training the stereo backbones from scratch to perform LiDAR-event stereo fusion. This allows either the deployment of strategies that process the LiDAR data directly as input \cite{cheng2019noise,wang20193d} or those not requiring it, i.e., \cite{poggi2019guided} and ours. Specifically, Concat \cite{cheng2019noise} and Guided+Concat \cite{wang20193d} strategy achieve results comparable to those by VSH and BTH observed before, thus outperforming Guided \cite{poggi2019guided} which, on the contrary, cannot benefit much from the training process.
When deploying our solutions during training, their effectiveness dramatically increases, often dropping 1PE error below 10\%. VSH often yields the best 1PE and MAE overall, nominating it as the most effective -- yet intrusive -- among our solutions. 

\begin{table*}[t]
    \centering
    \caption{\textbf{Results on M3ED \cite{Chaney_2023_CVPR} -- pre-trained.} We test the different stacked representations (rows) with several fusion strategies applied to pre-trained stereo backbones.}
    \renewcommand{\tabcolsep}{7pt}
    \scalebox{0.6}{
    \begin{tabular}{c}

    \hspace{-0.4cm}\begin{tabular}{crrr:rrr:rrr:rrr:rrr}
        \Xhline{3pt}
        & \multicolumn{3}{l;}{Stacked} & \multicolumn{3}{c:}{Baseline} & \multicolumn{3}{c:}{Guided \cite{poggi2019guided}} & \multicolumn{3}{c:}{VSH (ours)} & \multicolumn{3}{c}{BTH (ours)} \\
        & \multicolumn{3}{l;}{representation} & 1PE & 2PE & MAE & 1PE & 2PE & MAE & 1PE & 2PE & MAE & 1PE & 2PE & MAE \\
        \Xhline{3pt}
        (A) & \multicolumn{3}{l;}{Histogram \cite{maqueda2018event}} & 37.70 & 19.49 & 1.76 & 37.18 & 19.29 & 1.75 & \fst 20.19 & \fst 11.19 & \fst 1.19 & \snd 22.32 & \snd 12.37 & \snd 1.27 \\
        (B) & \multicolumn{3}{l;}{MDES \cite{nam2022stereo}} & 43.17 & 19.50 & 1.85 & 42.27 & 19.16 & 1.83 & \snd 29.42 & \snd 14.80 & \snd 1.52 & \fst 22.58 & \fst 12.20 & \fst 1.30 \\
        (C) & \multicolumn{3}{l;}{Concentration \cite{nam2022stereo}} & 45.78 & 20.84 & 1.82 & 45.06 & 20.57 & 1.80 & \snd 33.63 & \snd 16.19 & \snd 1.53 & \fst 25.22 & \fst 12.68 & \fst 1.28 \\
        (D) & \multicolumn{3}{l;}{Voxelgrid \cite{zhu2019unsupervised}} & 37.33 & 17.66 & 1.70 & 36.64 & 17.38 & 1.68 & \fst 20.40 & \fst 11.41 & \fst 1.22 & \snd 20.94 & \snd 11.72 & \snd 1.23 \\
        (E) & \multicolumn{3}{l;}{TORE \cite{baldwin2022time}} & 41.70 & 19.09 & 1.81 & 41.00 & 18.78 & 1.80 & \snd 28.25 & \snd 14.01 & \snd 1.47 & \fst 21.91 & \fst 12.34 & \fst 1.30 \\
        (F) & \multicolumn{3}{l;}{Time Surface \cite{lagorce2016hots}} & 38.58 & 18.52 & 1.72 & 37.91 & 18.23 & 1.70 & \snd 24.89 & \snd 13.34 & \snd 1.37 & \fst 22.60 & \fst 12.77 & \fst 1.31 \\
        (G) & \multicolumn{3}{l;}{ERGO-12 \cite{Zubic_2023_ICCV}} & 36.33 & 17.81 & 1.66 & 35.61 & 17.50 & 1.64 & \snd 22.53 & \snd 12.33 & \snd 1.26 & \fst 20.41 & \fst 11.69 & \fst 1.21 \\
        (H) & \multicolumn{3}{l;}{Tencode \cite{Huang_2023_WACV}} & 43.56 & 20.07 & 1.82 & 42.66 & 19.76 & 1.80 & \snd 28.24 & \snd 14.46 & \snd 1.43 & \fst 22.61 & \fst 12.75 & \fst 1.26 \\
        \Xhline{3pt}
        & \multicolumn{3}{l;}{Avg. Rank.} & \multicolumn{3}{c:}{-} & 3.00 & 3.00 & 3.00 & \snd 1.75 & \snd 1.75 & \snd 1.75 & \fst 1.25 & \fst 1.25 & \fst 1.25 \\
        \Xhline{3pt}
    \end{tabular}    
    \end{tabular}
    }
    
    \label{tab:m3ed_notrain}
\end{table*}
\begin{table*}[t]
    \centering
    \caption{\textbf{Results on M3ED \cite{Chaney_2023_CVPR} -- retrained.} We test different stacked representations (rows) with several fusion strategies applied during training.}
    \renewcommand{\tabcolsep}{7pt}
    \scalebox{0.6}{
    \begin{tabular}{c}

    \hspace{-0.4cm}\begin{tabular}{crrr:rrr:rrr:rrr:rrr}
        \Xhline{3pt}
        & \multicolumn{3}{c:}{Concat \cite{cheng2019noise}} & \multicolumn{3}{c:}{Guided+Concat \cite{wang20193d}} & \multicolumn{3}{c:}{Guided \cite{poggi2019guided}} & \multicolumn{3}{c:}{VSH (ours)} & \multicolumn{3}{c}{BTH (ours)} \\ 
        & 1PE & 2PE & \multicolumn{1}{r:}{MAE}  & 1PE & 2PE & \multicolumn{1}{r:}{MAE} & 1PE & 2PE & MAE & 1PE & 2PE & MAE & 1PE & 2PE & MAE \\
        \Xhline{3pt}
        (A) & 34.67 & 15.21 & 1.92 & 38.65 & 17.00 & 1.94 & 37.45 & 18.98 & 1.76 & \fst 19.34 & \fst 12.93 & \snd 1.46 & \snd 19.83 & \snd 13.20 & \fst 1.39 \\
        (B) & 37.72 & 16.91 & 1.85 & 37.32 & 17.16 & 2.14 & 37.00 & 18.66 & 1.76 & \snd 19.24 & \snd 13.17 & \snd 1.44 & \fst 18.70 & \fst 11.79 & \fst 1.24\\
        (C) & 39.88 & 19.01 & 2.33 & 38.45 & 17.76 & 2.47 & 38.14 & 19.62 & \snd 1.80 & \snd 19.84 & \snd 13.68 & 1.90 & \fst 19.46 & \fst 12.29 & \fst 1.35 \\
        (D) & 33.89 & 16.21 & 1.89 & 33.54 & 15.85 & 1.75 & 37.85 & 18.81 & \snd 1.74 & \fst 18.56 & \fst 11.76 & \fst 1.32 & \snd 21.02 & \snd 14.30 & 1.80\\
        (E) & 38.83 & 18.38 & 2.27 & 35.80 & 16.63 & 2.05 & 40.51 & 19.96 & 1.95 & \fst 20.03 & \snd 13.97 & \snd 1.86 & \snd 20.04 & \fst 12.65 & \fst 1.39\\
        (F) & 40.26 & 18.44 & 2.19 & 35.48 & 17.74 & 2.15 & 38.77 & 18.41 & 1.75 & \fst 19.61 & \fst 13.01 & \fst 1.55 & \snd 21.91 & \snd 14.33 & \snd 1.72 \\
        (G) & 42.43 & 19.31 & 2.31 & 42.24 & 18.42 & 2.34 & 37.95 & 17.83 & 1.76 & \fst 18.45 & \snd 12.31 & \snd 1.55 & \snd 19.12 & \fst 11.60 & \fst 1.24 \\
        (H) & 37.46 & 17.87 & 2.15 & 33.69 & 16.47 & 1.95 & 39.78 & 19.42 & 1.82 & \snd 19.49 & \snd 12.21 & \snd 1.38 & \fst 19.28 & \fst 11.68 & \fst 1.33 \\
        \Xhline{3pt}
          & 4.38 & 4.00 & 4.50 & 3.63 & 3.38 & 4.38 & 4.00 & 4.63 & 2.75 & \fst 1.38 & \snd 1.63 & \snd 1.88 & \snd 1.63 & \fst 1.38 & \fst 1.50 \\
        \Xhline{3pt}
    \end{tabular}    

    \end{tabular}
    }
    
    \label{tab:m3ed_train}
\end{table*}

\subsection{Experiments on M3ED}
\label{sec:experiments_m3ed}

We test the effectiveness of BTH and alternative approaches on M3ED, using the backbones trained on DSEC \textbf{without} any fine-tuning on M3ED itself.

\textbf{Pre-trained models.} \cref{tab:m3ed_notrain} collects the outcome of this experiment by applying Guided, VSH, and BTH to pre-trained models. 
Looking at the baselines, we can appreciate how M3ED is very challenging for models trained in a different domain, with 1PE errors higher than 30\%. This is caused by both the domain shift and the higher resolution of the event cameras used. Even so, complex event representations -- \eg, ERGO-12 \cite{Zubic_2023_ICCV} -- can better generalize. Guided confirms its limited impact, this time mainly because of the ineffectiveness of the cost volume modulation in the absence of any information from the events domain. On the contrary, we can appreciate even further the impact of VSH and BTH, almost halving the 1PE error. Specifically, BTH is the absolute winner with 6 out of 8 representations, and the best choice for pre-trained frameworks.

\begin{figure}[t]
    \centering
    \renewcommand{\tabcolsep}{1pt}
    \begin{tabular}{ccccc}

        \tiny Events \& LiDAR & \tiny Baseline & \tiny Guided \cite{poggi2019guided} & \tiny BTH (ours) & \tiny BTH (ours, retrain) \\ 
        \includegraphics[trim=0cm 2cm 0cm 0cm, clip, width=0.17\linewidth]{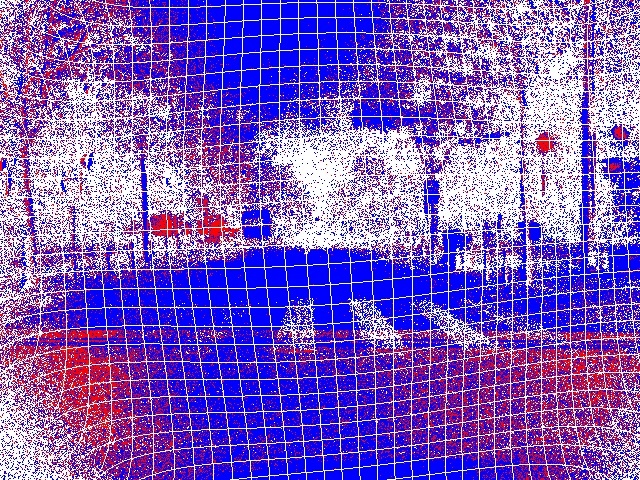} 
        & 
        \includegraphics[trim=0cm 2cm 0cm 0cm, clip, width=0.17\linewidth]{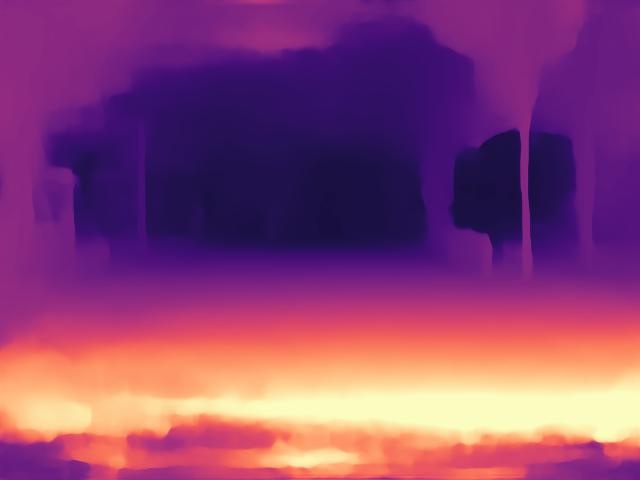} & 
         \includegraphics[trim=0cm 2cm 0cm 0cm, clip, width=0.17\linewidth]{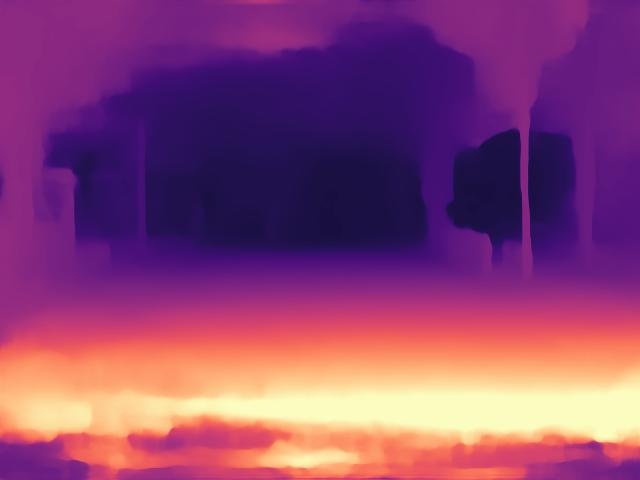} & 
         \includegraphics[trim=0cm 2cm 0cm 0cm, clip, width=0.17\linewidth]{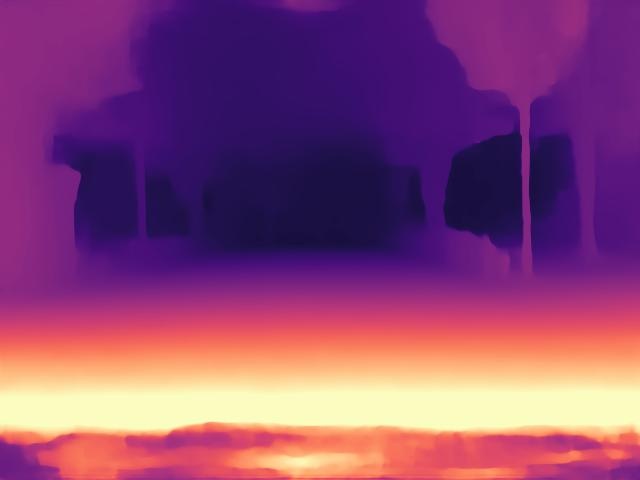} & 
         \includegraphics[trim=0cm 2cm 0cm 0cm, clip, width=0.17\linewidth]{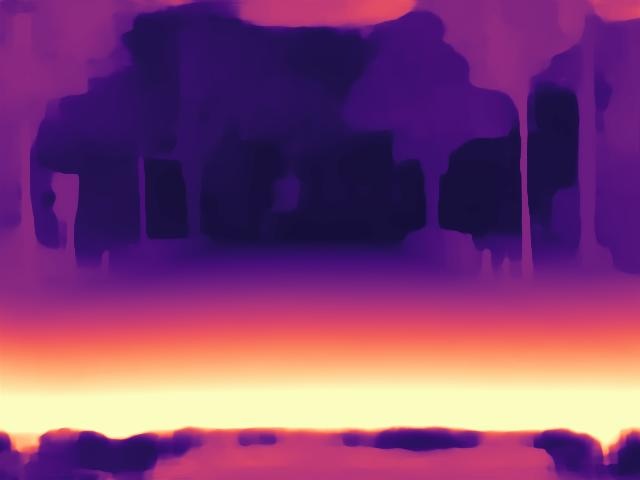} \\
        
        \begin{overpic}[trim=0cm 2cm 0cm 0cm, clip, width=0.17\linewidth]{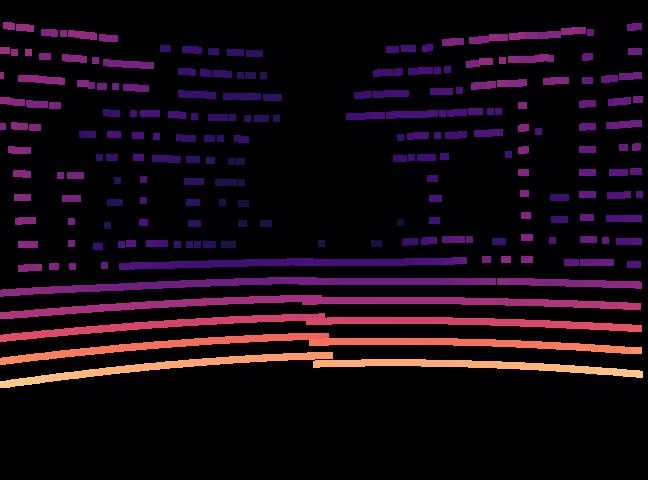}\end{overpic} &
        \begin{overpic}
        [trim=0cm 2cm 0cm 0cm, clip, width=0.17\linewidth]{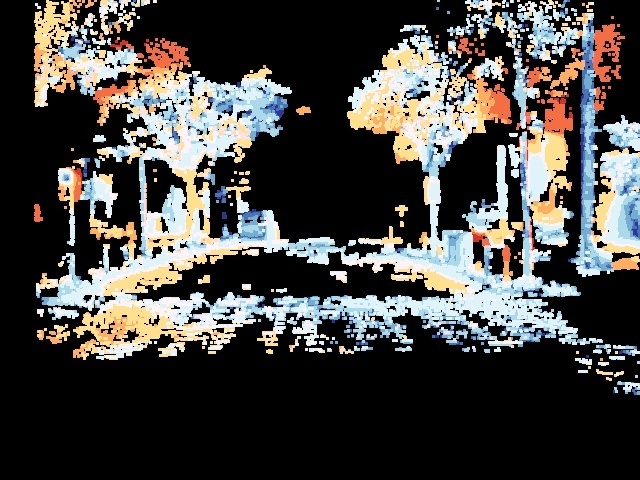}
        \put(0,2){
        \tiny\textcolor{white}{\textbf{1PE: 50.84\%}}}
        \end{overpic} &
         \begin{overpic}
         [trim=0cm 2cm 0cm 0cm, clip, width=0.17\linewidth]{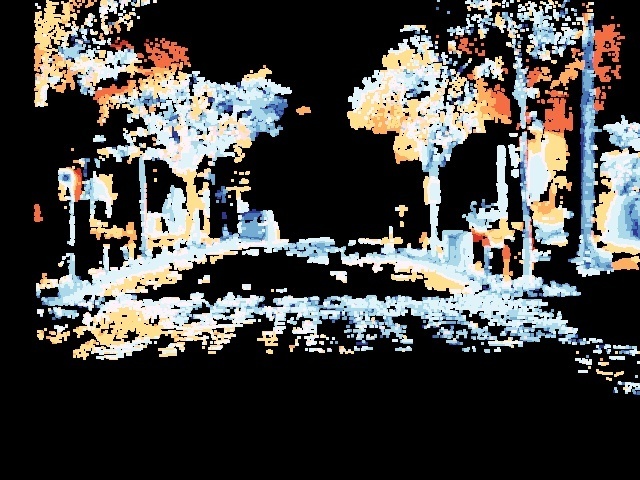}
         \put(0,2){
        \tiny\textcolor{white}{\textbf{1PE: 48.94\%}}}
         \end{overpic} & 
         \begin{overpic}[trim=0cm 2cm 0cm 0cm, clip, width=0.17\linewidth]{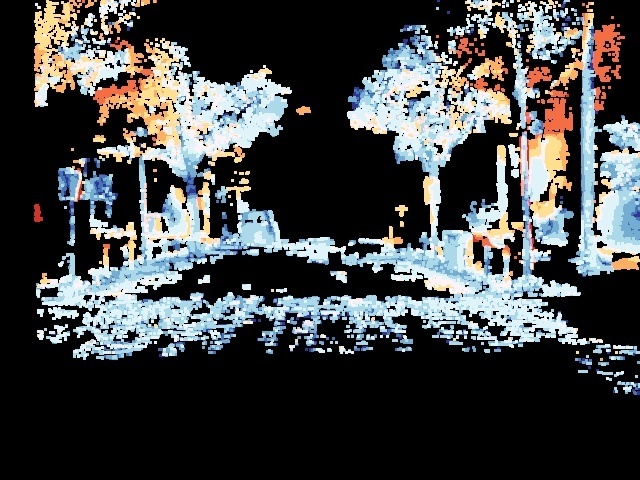} 
         \put(0,2){
        \tiny\textcolor{white}{\textbf{1PE: 32.16\%}}}
         \end{overpic} &
         \begin{overpic}[trim=0cm 2cm 0cm 0cm, clip, width=0.17\linewidth]{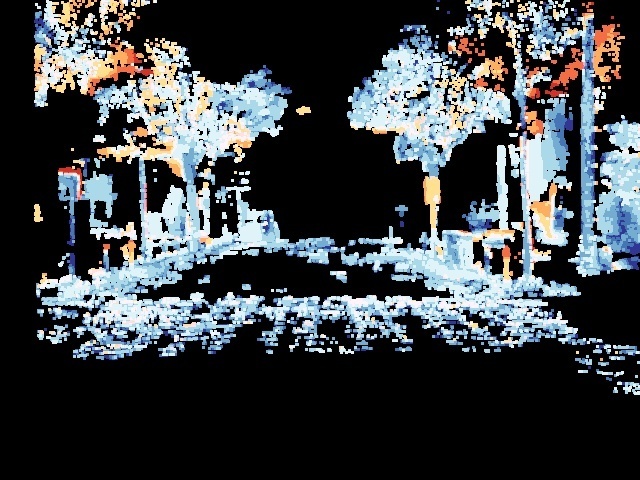} 
         \put(0,2){
        \tiny\textcolor{white}{\textbf{1PE: 22.83\%}}}
         \end{overpic}\vspace{0.1cm} \\

         \tiny Events \& LiDAR & \tiny Baseline & \tiny Guided \cite{poggi2019guided} & \tiny VSH (ours) & \tiny BTH (ours) \\ 
        \includegraphics[width=0.17\linewidth]{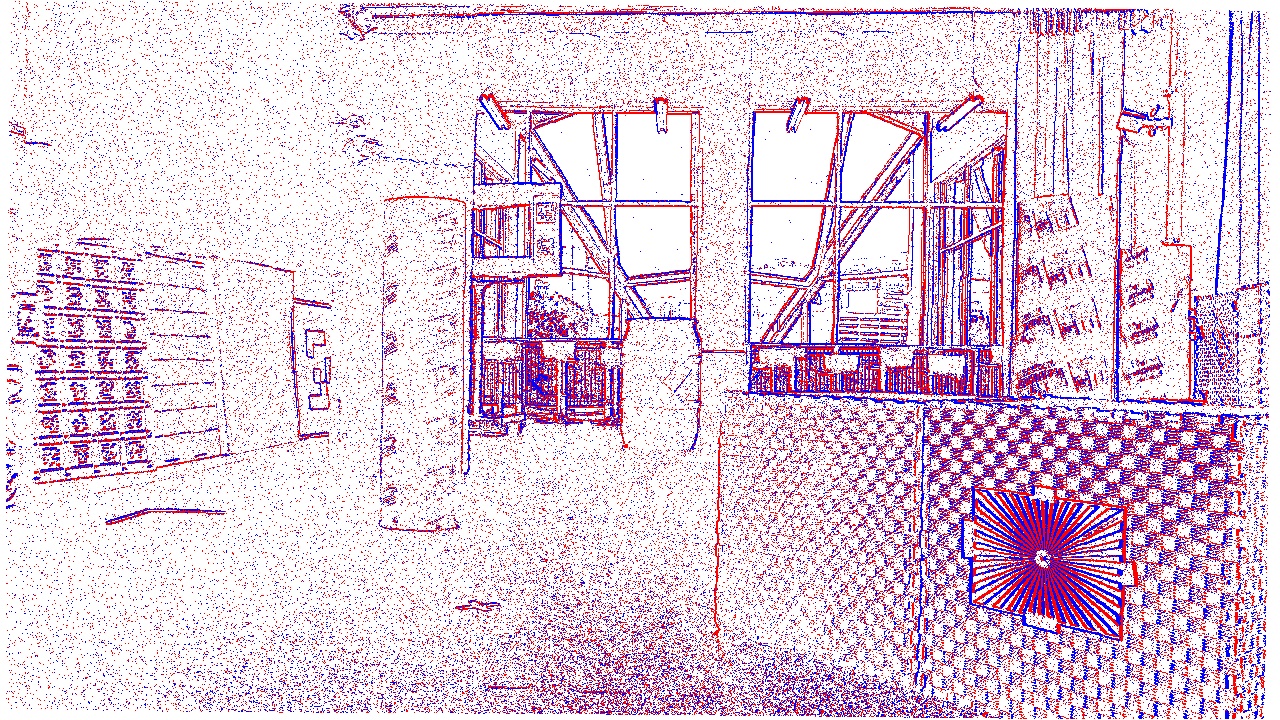} 
        & 
        \includegraphics[width=0.17\linewidth]{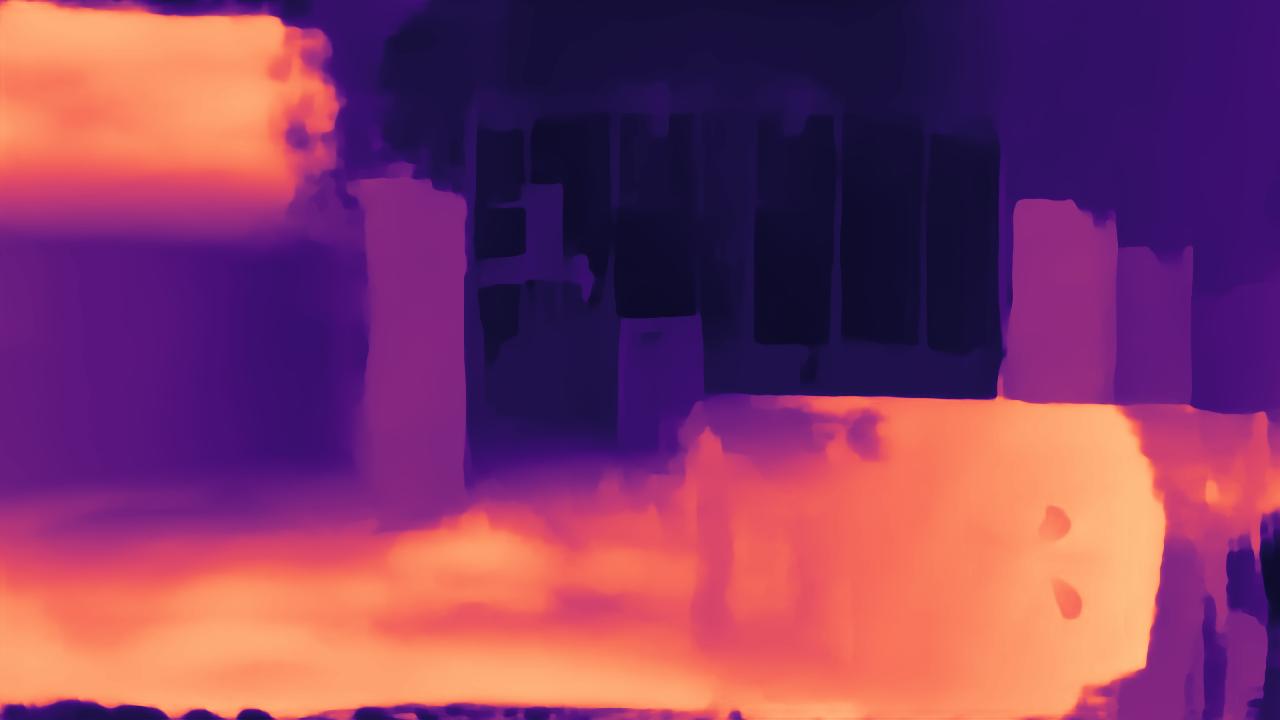} & 
         \includegraphics[width=0.17\linewidth]{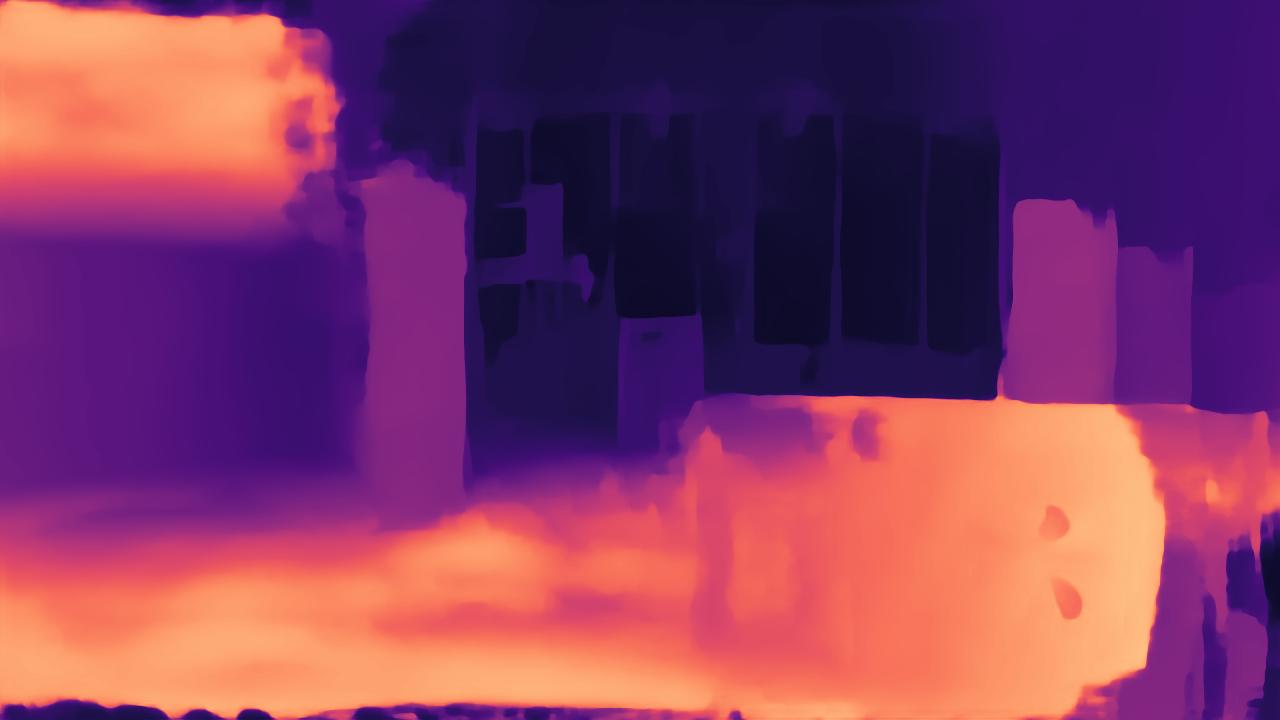} & 
         \includegraphics[width=0.17\linewidth]{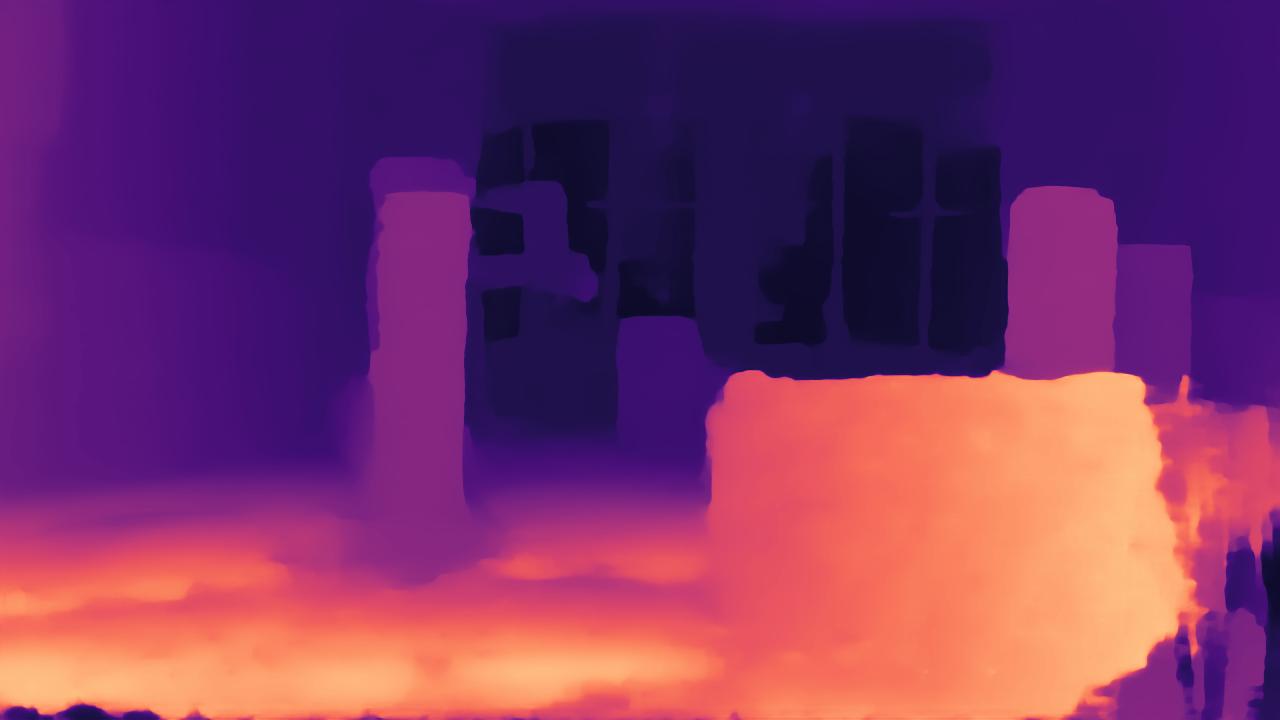} & 
         \includegraphics[width=0.17\linewidth]{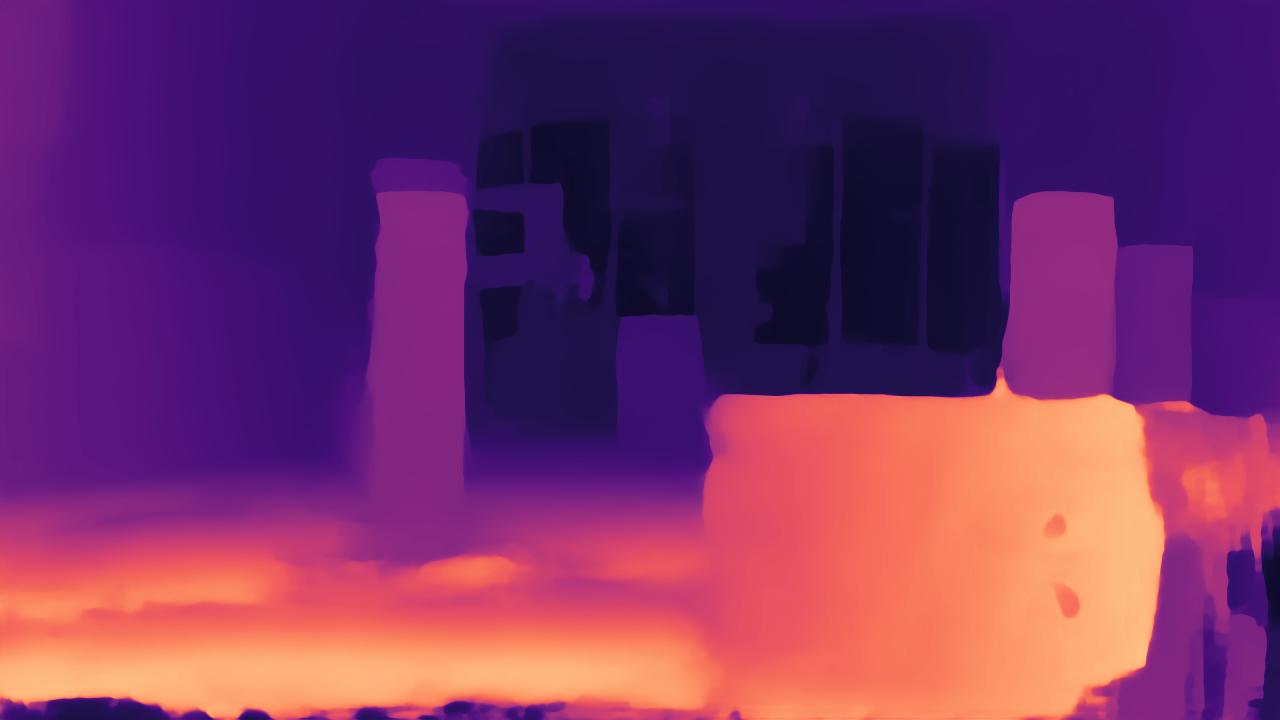} \\
        
        \includegraphics[width=0.17\linewidth]{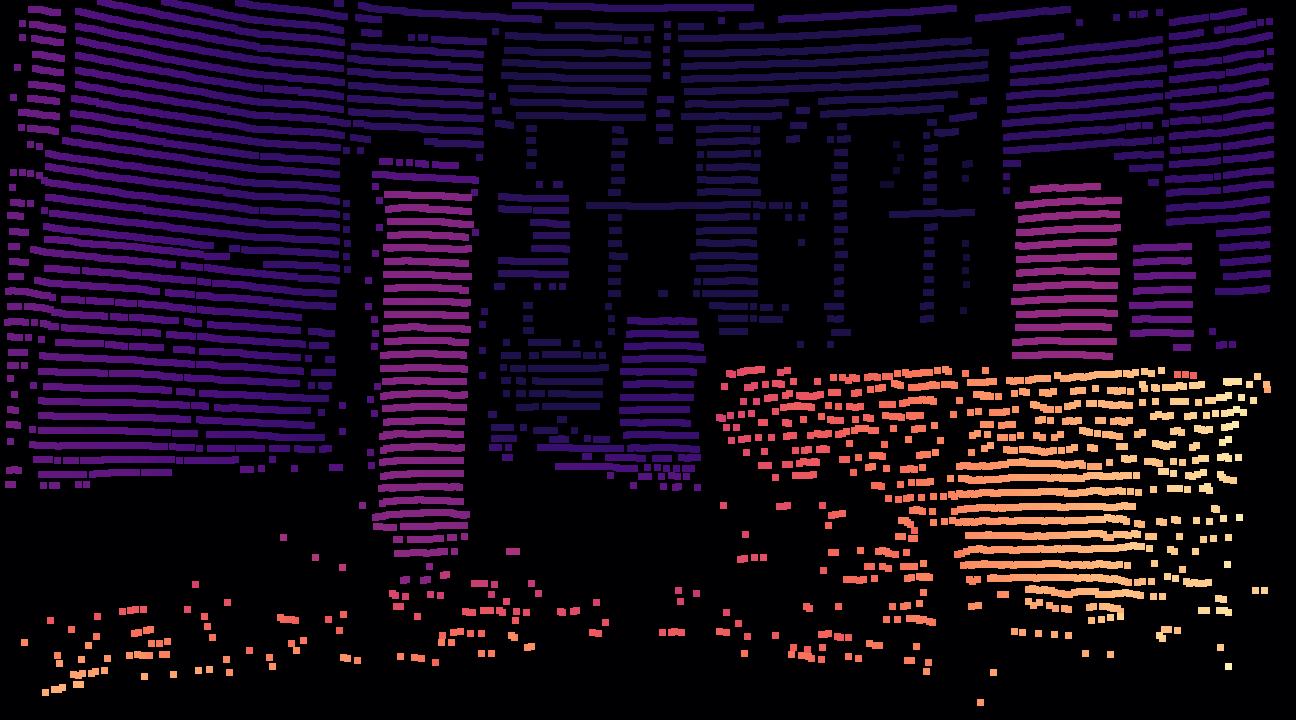} &
        \begin{overpic}
        [width=0.17\linewidth]{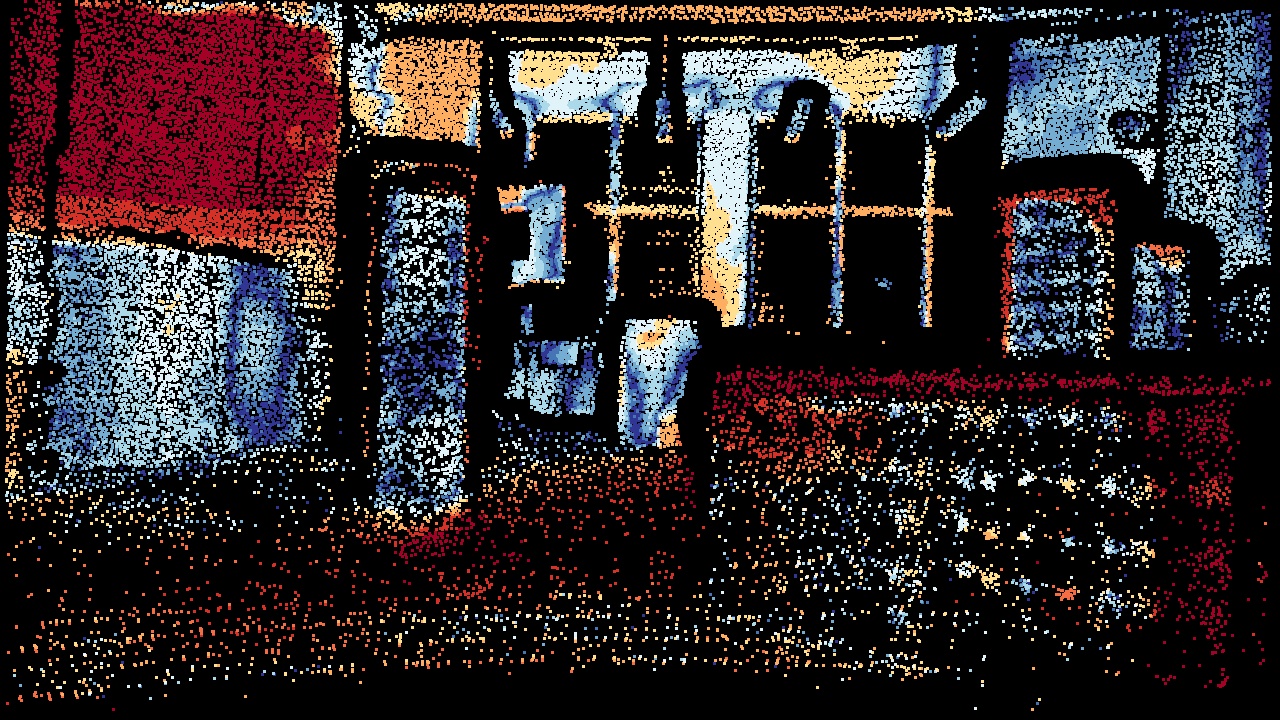}
        \put(0,2){
        \tiny\textcolor{white}{\textbf{1PE: 53.19\%}}}
        \end{overpic} &
         \begin{overpic}
         [width=0.17\linewidth]{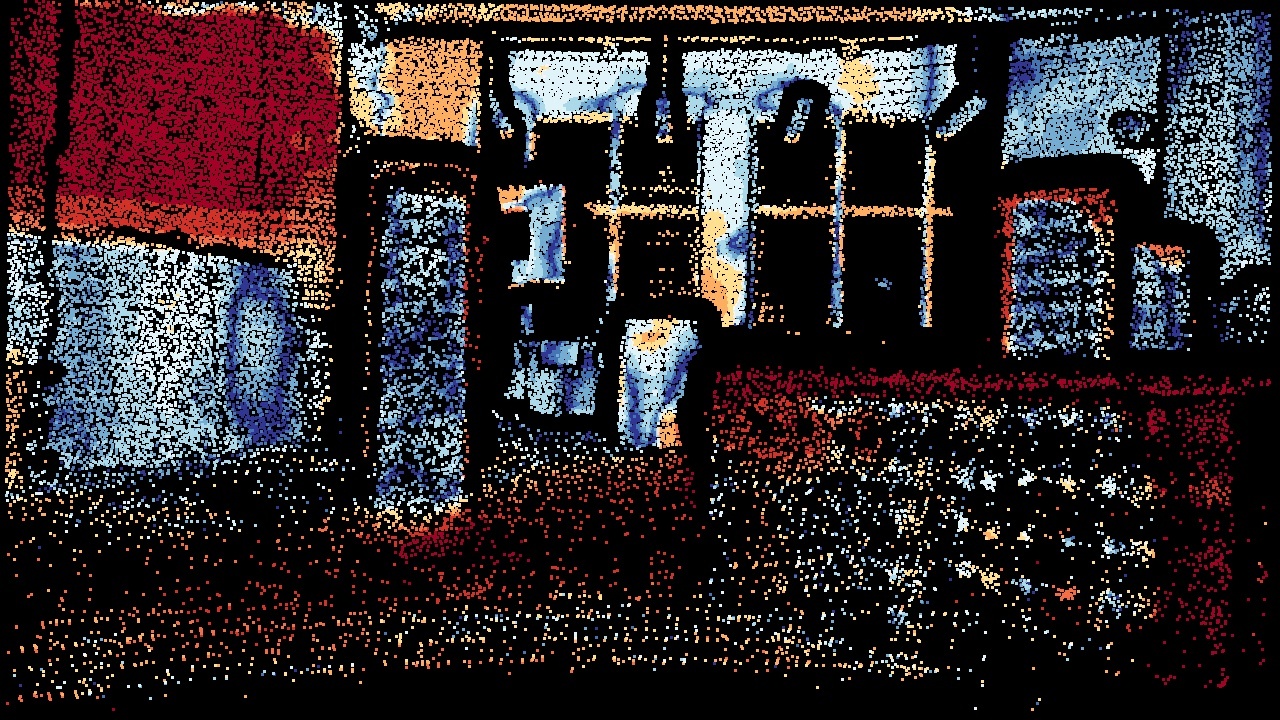}
         \put(0,2){
        \tiny\textcolor{white}{\textbf{1PE: 50.26\%}}}
         \end{overpic} & 
         \begin{overpic}[width=0.17\linewidth]{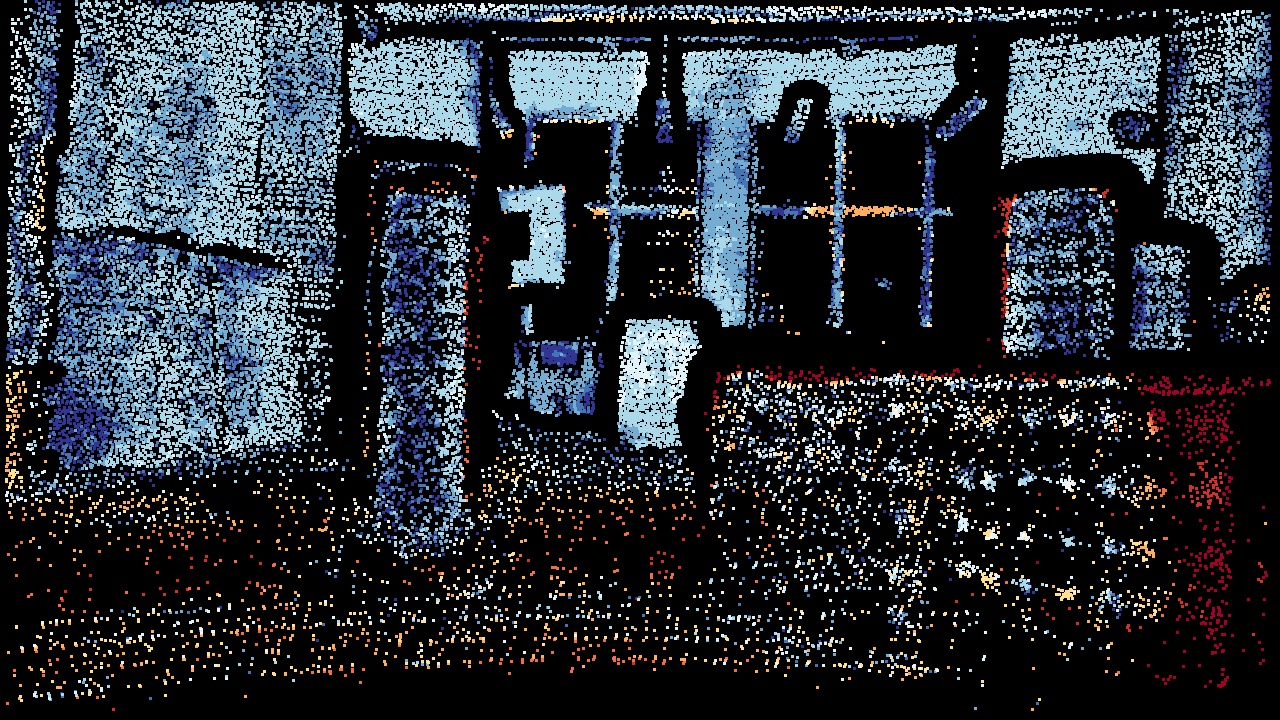} 
         \put(0,2){
        \tiny\textcolor{white}{\textbf{1PE: 10.94\%}}}
         \end{overpic} &
         \begin{overpic}[width=0.17\linewidth]{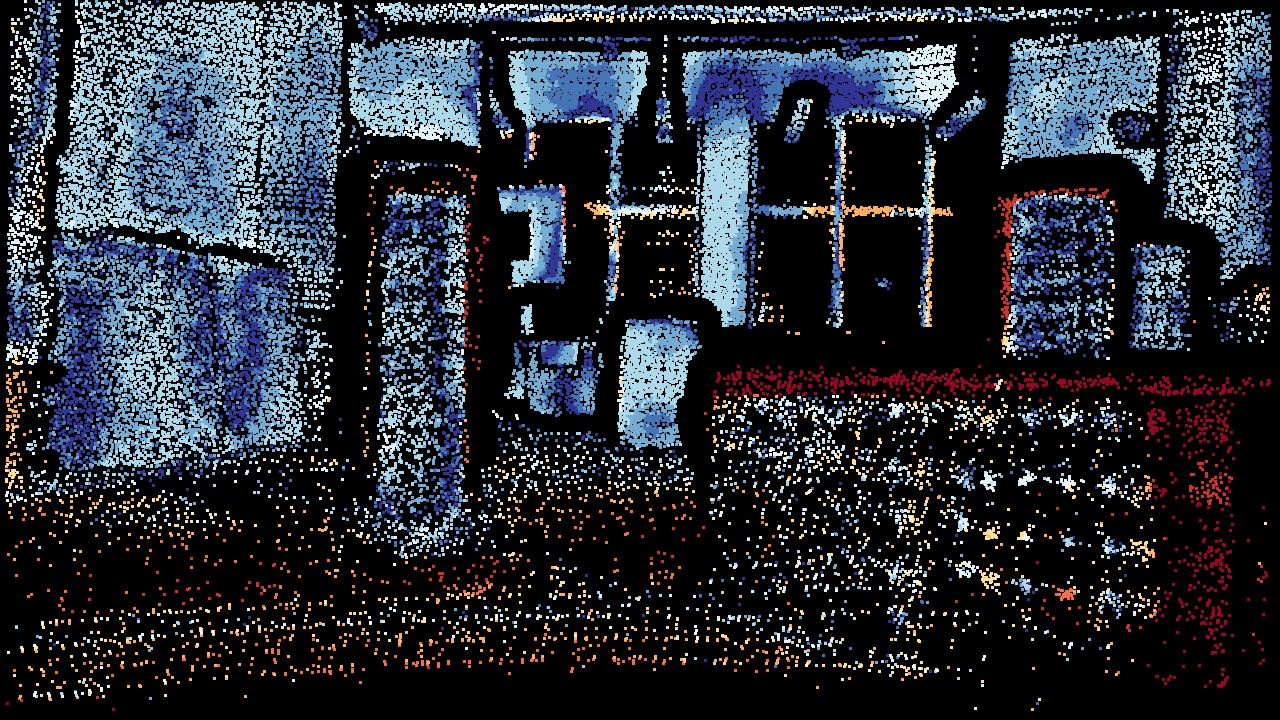} 
         \put(0,2){
        \tiny\textcolor{white}{\textbf{1PE: 13.50\%}}}
         \end{overpic}\vspace{0.1cm} \\
    \end{tabular}
    \caption{\textbf{Qualitative results.} Results on DSEC \textit{zurich\_10\_b} with Voxelgrid \cite{zhu2019unsupervised} (top) and M3ED \textit{spot\_indoor\_obstacles} with Histogram \cite{maqueda2018event} (bottom).}
    \label{fig:qualitative}
\end{figure}

\textbf{Training from scratch.} \cref{tab:m3ed_train} resumes the results obtained when training on DSEC the backbones implementing LiDAR-event stereo fusion strategies. The very different distribution of depth points observed across the two datasets -- sourced respectively from 16 and 64-line LiDARs -- yields mixed results for existing methods \cite{cheng2019noise,wang20193d,poggi2019guided}, with rare cases for which they fail to improve the baseline model (e.g., Concat and Guided with Time Surface and ERGO-12, Guided+Concat with Histogram). On the contrary, backbones trained with VSH and BTH consistently improve over the baseline, often with larger gains compared to their use with pre-trained models. Overall, BTH is the best on 2PE and MAE, confirming it is better suited for robustness across domains and different LiDAR sensors.

\cref{fig:qualitative} shows qualitative results. On DSEC (top), BTH dramatically improves results over the baseline and Guided, yet cannot fully recover some details in the scene except when retraining the stereo backbone. On M3ED (bottom), both VSH and BTH with pre-trained models reduce the error by 5$\times$.

\subsection{Experiments on M3ED -- Time-misaligned LiDAR} 
\label{sec:m3ed_offset}

We conclude by assessing the robustness of the considered strategies against the use of LiDAR not synchronized with the timestamp at which we wish to estimate disparity -- occurring if we wish to maintain the microsecond resolution of the event cameras.
Purposely, we extract raw LiDAR measurements collected 3, 13, 32, 61, and 100 ms in the past with the M3ED SDK.

\begin{figure*}[t]
    \centering
    \includegraphics[trim=0cm 0cm 0cm 0cm, clip,width=\linewidth]{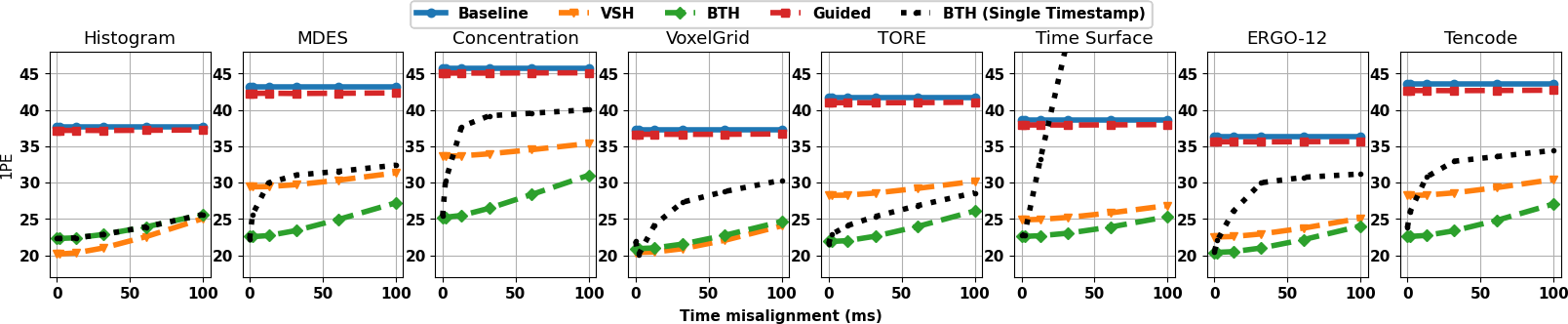}
    \caption{\textbf{Experiments with time-misaligned LiDAR on M3ED \cite{Chaney_2023_CVPR} -- pre-trained.} We measure the robustness of different fusion strategies against the use of out-of-sync LiDAR data, without retraining the stereo backbone.}
    \label{fig:offset_notrain}
\end{figure*}

\begin{figure*}[t]
    \centering
    \includegraphics[trim=0cm 0cm 0cm 0cm, clip,width=\linewidth]{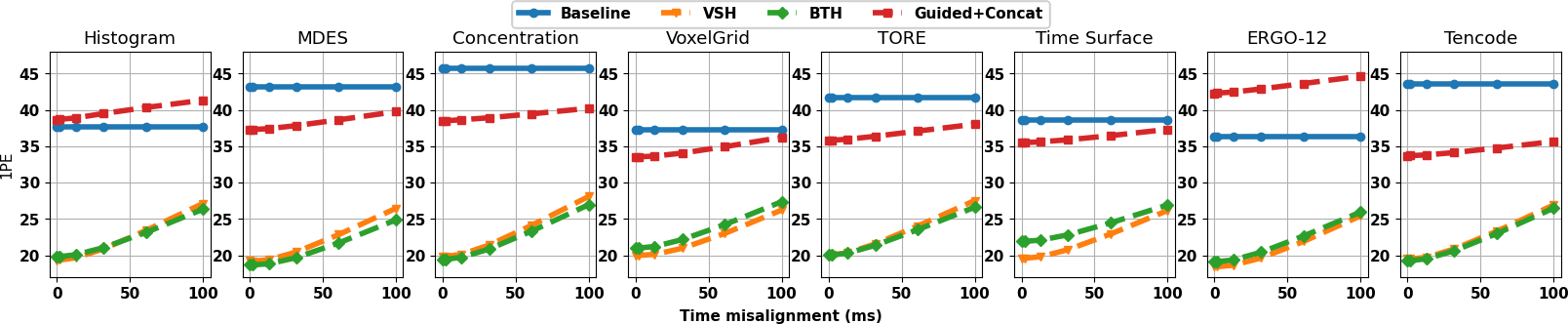}
    \caption{\textbf{Experiments with time-misaligned LiDAR on M3ED \cite{Chaney_2023_CVPR} -- retrained.} We measure the robustness of different fusion strategies against the use of out-of-sync LiDAR data when training the stereo backbone from scratch.}
    \label{fig:offset_train}
\end{figure*}

\cref{fig:offset_notrain} shows the trend of the 1PE metric achieved by Guided (red), VSH (yellow) and BTH (black and green) on pre-trained backbones. Not surprisingly, the error rates arise at the increase of the temporal distance: while this is less evident with Guided because of its limited impact, this becomes clear with VSH and BTH. Nonetheless, both can always retain a significant gain over the baseline model (blue) -- i.e., the stereo backbone processing events only -- even with the farthest possible misalignment with a 10Hz LiDAR (100ms). We can appreciate how BTH is often better than VSH (coherently with \cref{tab:m3ed_notrain}), yet only when repeated injections are performed (green). Indeed, using a single injection (black) rapidly leads BTH to an accuracy drop when increasing the misalignment, except when using Histogram representation. Overall, BTH with ERGO-12 is the most robust solution.
\cref{fig:offset_train} shows the results achieved by VSH (yellow) and BTH (green) after retraining, against the best competitor according to average ranks in \cref{tab:m3ed_train} -- i.e., Guided+Concat (red). The impact of this latter is limited and sometimes fails to improve the baseline (see Histogram and ERGO-12). On the contrary, our solutions confirm their robustness and effectiveness even when dealing with time-misaligned LiDAR data.

\section{Conclusion}
\label{sec:conclusion}

This paper proposes a novel framework for implementing event stereo and LiDAR fusion. It works by hallucinating fictitious events either in the stacked representation processed by stereo backbones or the continuous streams sensed by event cameras, easing the matching process to the downstream stereo model estimating disparity. Our exhaustive experiments prove that our solutions, VSH and BTH, dramatically outperform alternative fusion strategies from the RGB stereo literature, retaining the microsecond resolution typical of event cameras despite the discrete frame rate of LiDARs and depth sensors in general.

\textbf{Limitations.} Despite the robustness shown with misaligned LiDAR data, a marginal drop in accuracy compared to the case of having LiDAR measurements at the very same timestamp at which we aim to infer disparity maps occurs. Future work will focus on studying new design mechanisms to deal with it.

{\textbf{Acknowledgement.} 
This study was carried out within the MOST – Sustainable Mobility National Research Center and received funding from the European Union Next-GenerationEU – PIANO NAZIONALE DI RIPRESA E RESILIENZA (PNRR) – MISSIONE 4 COMPONENTE 2, INVESTIMENTO 1.4 – D.D. 1033 17/06/2022, CN00000023. This manuscript reflects only the authors’ views and opinions, neither the European Union nor the European Commission can be considered responsible for them.

We acknowledge the CINECA award under the ISCRA initiative, for the availability of high-performance computing resources and support.}

\bibliographystyle{splncs04}
\bibliography{main}

\phantom{Supplementary}
\multido{\i=1+1}{13}{


\section*{Supplementary material}

# LiDAR-Event Stereo Fusion with Hallucinations

## Supplementary Material

This document provides additional details regarding ECCV 2024 paper "LiDAR-Event Stereo Fusion with Hallucinations". Specifically, we report:

- **page 21:** the impact of our framework on event streams distinctiveness
- **pages 22-25:** the composition of the datasets split used in our experiments, as well as the pre-processing pipeline necessary to obtain data suitable for our purposes from both DSEC [7] and M3ED [5]
- **pages 26-27:** detailed description of the LiDAR-stereo fusion strategies inherited from classical deep stereo literature [6, 13, 16]
- **page 28:** additional ablation studies
- **pages 29-32:** more qualitative results on the M3ED dataset [5]

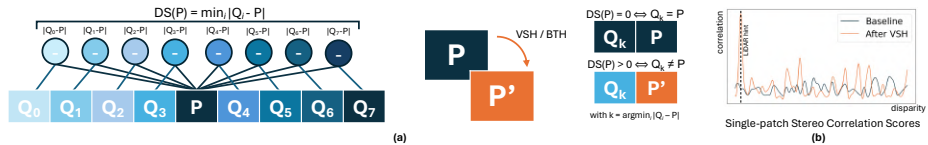

Fig. I: Distinctiveness before/after VSH/BTH (a), correlation scores (b).

## 1 Impact on Distinctiveness

In this section, we discuss the reasons behind the effectiveness of our hallucination strategies from a probabilistic perspective. In the stereo matching tasks, ambiguities arise when the patches to match are not distinctive – i.e., multiple patches along the horizontal epipolar line are identical. For a single patch $P$ and its neighbors $Q_i$, we can define its **distinctiveness** [10] $DS(P)$ as $\min_i |P - Q_i|$. Such distinctiveness is 0 if at least one $Q_i$ is identical to $P$ – see Fig. I (a). This is very likely to occur for patches for which no event at all is triggered by the camera, as for the $\approx 30\%$ of the patches on the M3ED dataset.

As VSH puts a random $P'$ on both left and right frames, it will be sufficient for $P'$ to have $DS(P') > 0$ to ease the matching of those empty patches. A specific $P'$ is generated with probability $p(P') = \frac{1}{V^{N^2 \times B}}$, being $V$ the possible per-pixel values (e.g., $2^8$ for uint8), with patch size $N \times N$ and $B$ stack channels. Accordingly, the probability $p(DS(P') = 0)$ equals the probability of having another patch on the horizontal scanline identical to $P'$, i.e., $W \times p(P')$ if we assume patches to be independent (in the worst case scenario), being $W$ the image width. On M3ED, $p(DS(P') = 0) = 1280 \times p(P)$, that is $\approx 5e^{-41}$ and $\approx e^{-257}$ for Histogram and Voxel Grid respectively. This means VSH has probability $\approx 1$ to ease matching for empty patches if a LiDAR hint is available for them. We confirm this expected behavior in Fig. I (b), showing the correlation curve computed by the stereo network on an empty patch, peaked on the LiDAR value after VSH. Similar derivation can be done for BTH.

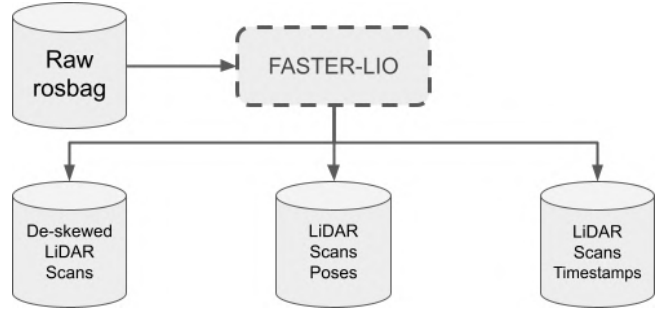

**Fig. II:** DSEC [7] processing scheme (1).

## 2  Evaluation Datasets and Pre-processing

In this section, we provide further details concerning the datasets used in our work and the pre-processing we carried out. In particular, we describe the DSEC [7] search split we used for the hyperparameters search concerning our proposals VSH and BTH, the M3ED [5] evaluation split we selected, and how we managed to process both datasets to extract raw LiDAR and, on M3ED dataset [5], for obtaining misaligned LiDAR measurements with respect to the timestamp at which we estimate disparity maps – and thus, at which we have ground-truth depth for evaluation.

### 2.1  DSEC [7] Dataset

We start with DSEC [7], which we use for i) tuning the hyper-parameters in our solutions, and ii) training the models involved in our experiments.

**Search split.** We select three sequences from the training set: *zurich_city_00_b*, *interlaken_00_c* and *zurich_city_09_c*.

**Processing Scheme.** We managed to extract raw LiDAR disparity maps directly from the rosbag files provided by the authors. Our extraction pipeline is detailed in Figs. II and III. Differently from M3ED [5], this dataset does not provide any ground-truth pose. Consequently, as the first step detailed in Fig. II, we deployed a LiDAR inertial odometry framework – FASTER-LIO [2] – to obtain point clouds without rolling shutter effects (*de-skewed*), their corresponding poses and timestamps.

Assuming a shared clock between ground-truth and raw timestamps, we aim to synchronize each ground-truth disparity map at time $t_d$ with the nearest de-skewed LiDAR scan $\mathbf{M}_d$. As shown in Fig. III, we can achieve this goal using a mapping function that links raw with ground-truth data using timestamps. Next, we can use the estimated poses to align $\mathbf{M}_d$ to its correlated ground-truth frame, using linear interpolation. Unfortunately, we empirically noticed that this step is insufficient to guarantee an acceptable alignment. To further refine this alignment, we deployed point-to-plane ICP algorithm [14] with L1 robust kernel [1]

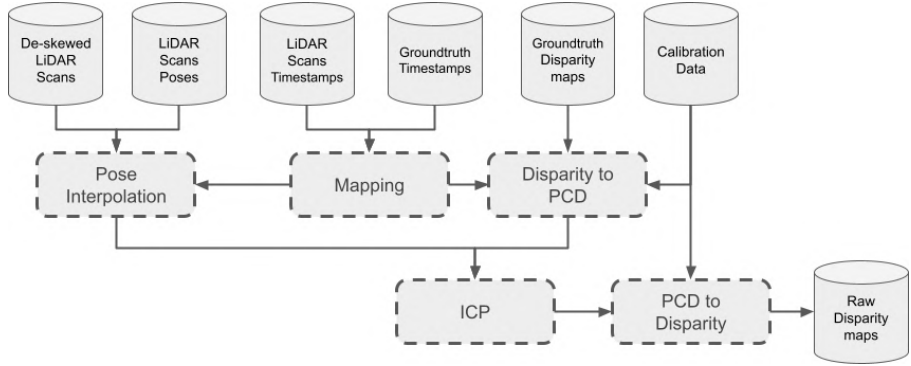

**Fig. III:** DSEC [7] processing scheme (2).

between $\mathbf{M}_d$ and the ground-truth point cloud (PCD) obtained by reprojection of the ground-truth disparity map to 3D. To ensure the best alignment possible, we repeated the process described in Fig. III also using the previous and the next LiDAR scans $\mathbf{M}_{d-1}, \mathbf{M}_{d+1}$ and skipping the pose interpolation sub-step. As a result, for each ground-truth frame we have six possible candidates – *i.e.*, $\{(\mathbf{M}_{d-1}, \mathbf{I}), (\mathbf{M}_{d-1}, \mathbf{P}_{d-1}), (\mathbf{M}_d, \mathbf{I}), (\mathbf{M}_d, \mathbf{P}_d), (\mathbf{M}_{d+1}, \mathbf{I}), (\mathbf{M}_{d+1}, \mathbf{P}_{d+1})\}$, where $\mathbf{I}$ (identity rigid transformation) and $\mathbf{P}_{d-1}, \mathbf{P}_d, \mathbf{P}_{d+1}$ (interpolated rigid transformation respectively for ground-truth frame $d-1, d, d+1$) represent the initial state for ICP algorithm. We select the one that best reduces the mean reprojection error (MAE). As the last step, raw point clouds are roto-translated into the left rectified event camera point of view and then projected into the rectified image plane to obtain depth maps, that are converted into disparity maps. Finally, we measure the MAE between each raw LiDAR depth map and its corresponding ground-truth depth map and discard those with an error exceeding 0.5 meters from training/search/testing splits.

### 2.2 M3ED [5] Dataset.

In this second dataset, we evaluate the generalization performance by any of the models involved in our evaluation, as well as we measure the robustness of LiDAR-event stereo frameworks when processing misaligned LiDAR data.

**Evaluation Split.** We select both some outdoor (*car_forest_tree_tunnel* and *car_urban_day_penno_small_loop*, with the former featuring a fully-static set of frames) and indoor (*falcon_indoor_flight_1*, *falcon_indoor_flight_2* and *falcon_indoor_flight_3*) sequences.

**Processing Scheme.** Figs. IV to VI sketch the main stage of our preprocessing pipeline. For each M3ED sequence, the authors provide *data* file and *depth_gt* file. The former provides raw LiDAR measurements, left and right distorted, unrectified stereo events, calibration parameters for each sensor, and other data not meaningful for our scope. The latter store ground-truth depth maps aligned to the left event camera and ground-truth poses for LiDAR (*i.e.*,

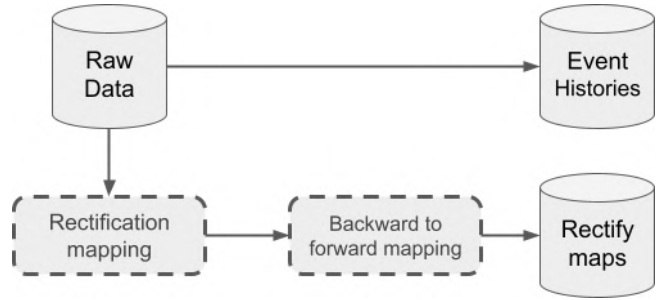

**Fig. IV: M3ED [5] processing scheme (1).** Extraction of stereo histories and rectification maps from the *data* file.

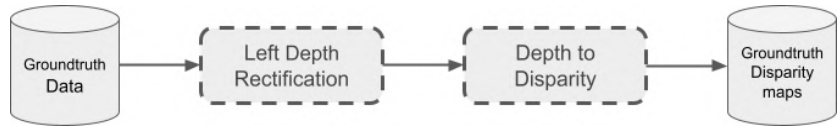

**Fig. V: M3ED [5] processing scheme (2).** After rectification, ground-truth depth maps are converted to disparity maps.

$Ln\_T\_L0$) and left event camera (*i.e.*, $Cn\_T\_C0$). This eases the LiDAR to ground-truth alignment with respect to DSEC dataset. All sensors, ground-truth depths, and poses are time-synchronized using a global clock. For more details, please refer to the original paper [5].

We extract the stereo event histories directly from the *data* file, as displayed in Fig. IV. Undistortion and rectification are managed at run-time using a forward mapping, stored in a single *rectify_map* file for each event camera. To create those mappings, we fed intrinsic, extrinsic, and distortion coefficients to the OpenCV functions. Furthermore, since OpenCV returns backward mapping functions, we used an iterative strategy to obtain left and right forward mappings. We maintain intermediate and final products of this step in memory, as they will be required by subsequent steps.

After the previous step, we process raw ground-truth depth maps from the *depth_gt* file (Fig. V). We rectify depth maps using the same left-forward mapping obtained in the previous step. For each depth map, we assume $t_d = t_z$ – *i.e.*, the timestamp $t_d$ at which we want to estimate disparity is the same as the timestamp $t_z$ at which the depth map is captured. Finally, depth maps are converted into disparity maps using stereo camera parameters obtained previously from stereo calibration.

The final step consists of processing raw LiDAR scans from the *data* file (Fig. VI). For each LiDAR scan, we linearly interpolate the pose at which the scan starts and the pose at which the scan ends as we need them to obtain a point cloud without rolling shutter effects (*de-skewed*). We aim to acquire both in-sync LiDAR disparity maps (*i.e.*, $t_d = t_z, \Delta t_o = 0$) and out-of-sync (or misaligned)

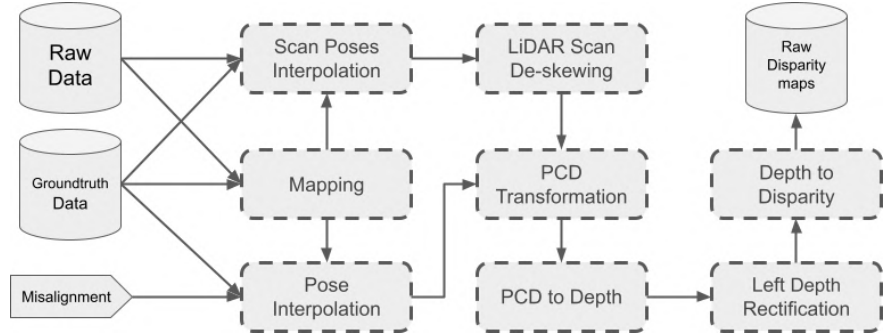

**Fig. VI: M3ED [5] processing scheme (3).** We used interpolated poses to correct raw LiDAR point clouds affected by rolling shutter effects. Then point clouds are roto-translated, projected into image plane, rectified and converted into disparity maps.

LiDAR disparity maps – *i.e.*, $t'_z = t_d - \Delta t_o, \Delta t_o > 0$, where $\Delta t_o$ is the temporal misalignment chosen arbitrarily (*e.g.*, 3, 13, 32, 61 and 100 ms) – these latter are used for experiments in Section 5.6. Given $\mathbf{M}_d$ as the closest de-skewed LiDAR scan to $t_d$ (mapping step in Fig. VI) and the temporal misalignment $\Delta t_o$, the former goal is achieved by roto-translating $\mathbf{M}_d$ as if it had been captured at $t'_z = t_d - \Delta t_o$. After that, we linearly interpolate the pose at time $t'_z$ using the two temporally nearest LiDAR poses (*i.e.*, $Ln\_T\_L0$). Next, point clouds are roto-translated into the left event camera point of view, and then projected into the image plane to obtain depth maps. Finally, depth maps are rectified and converted into disparity maps. Before the evaluation, we calculate the MAE between each raw LiDAR depth map and its corresponding ground-truth depth map and discard them if the error exceeds 0.1 meters, as well as discard the first and last 50 frames where raw LiDAR data fully overlaps with ground-truth.

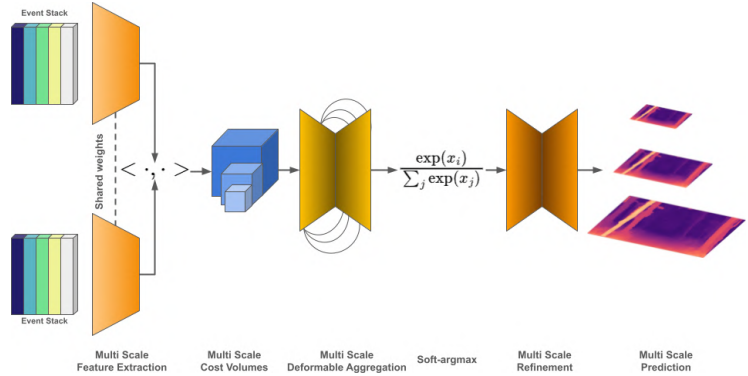

**Fig. VII: SE-CFF architecture.**

## 3 LiDAR-Stereo Matching Fusion Architectures

We report additional details about the deep architectures used in our experiments, starting from the baseline network, SE-CFF [12], and then showing the fusion strategies ported from classical deep stereo literature [6, 13, 15]

**Baseline Model: SE-CFF.** Fig. VII provides an overview of the SE-CFF [12] architecture assumed as the baseline in our work: i) rectified event streams are organized into MDES representations and then processed by a concentration network, producing single-channel stacks; ii) these are processed by two features extractor with shared weights, producing outputs respectively at $\frac{1}{3}, \frac{1}{6}, \frac{1}{12}$ of the original resolution; iii) these are used to build a multi-scale cost-volume computing the correlation between left and right features along the epipolar line; iv) a multi-scale, 2D network made with deformable convolutions is used to refine the cost volumes; v) an initial disparity map is obtained through a soft-argmax operator; vi) a multi-scale refinement network produces a set of refined disparity maps. We deploy eight variants of this architecture, one for each stacked representation considered in our experiments – i.e., stage i) is replaced with the different representations.

**Guided Stereo [13].** The first among the strategies inherited from classical deep stereo literature is depicted in Fig. VIII. Specifically, LiDAR data are used to modulate any of the multi-scale cost volumes $\mathcal{F}$, according to a Gaussian function

$$\mathcal{G} = \left(1 - v_{ij} + v_{ij} \cdot k \cdot e^{-\frac{(d-g_{ij})^2}{2c^2}}\right) \cdot \mathcal{F} \qquad (1)$$

with $k, c$ being the height and width of the Gaussian. Following [13], LiDAR points are downsampled through nearest-neighbor interpolation to act at the different resolutions.

**Concat [16].** An alternative strategy consists of providing the raw LiDAR data as an input to the stereo model. Fig. IX gives an overview of this lat-

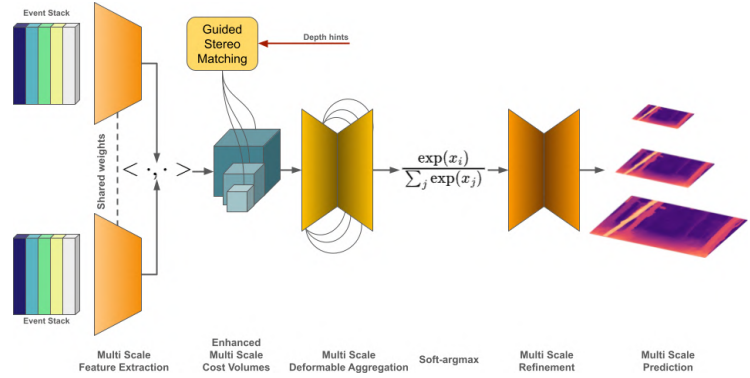

**Fig. VIII: Guided Stereo Matching [13] Framework.**

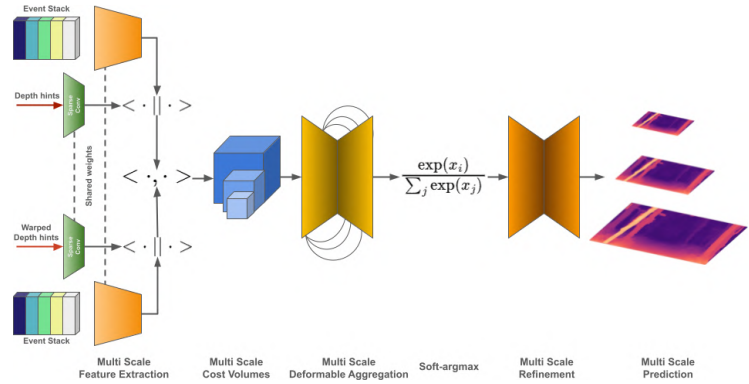

**Fig. IX: Concat [6] architecture.**

ter approach. LiDAR data are projected on left and right camera frames and processed by two feature extractors with shared-weights and made of Sparsity-Invariant Convolutions. These features are concatenated to those extracted from images – or stacked events, in our case – before the cost volumes are computed.

**Guided+Concat [6].** This strategy is inspired by CCVNorm architecture [6] and re-adapted to be deployed with SE-CFF. Specifically, CCVNorm deploys LiDAR data both as the input to the model, as well as to modulate the cost-volume at different stages during inference. Fig. X shows how we tailor this strategy to SE-CFF, basically by combining the two previous strategies.

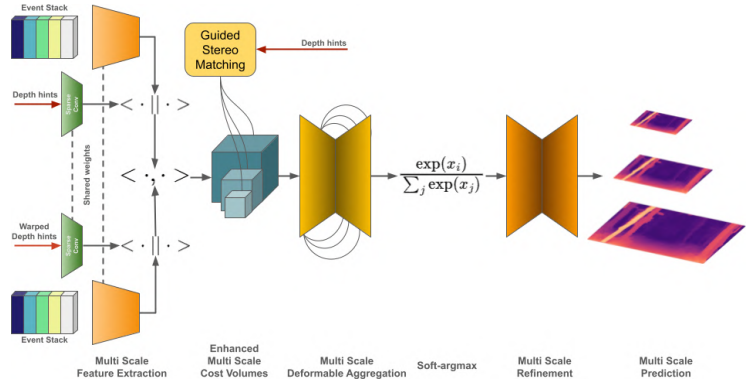

Fig. X: **Guided+Concat [15] architecture.**

**Table I: Ablation on VSH (top) and BTH (bottom).** We report 1PE for stereo backbones with different stacked representations on the DSEC search split.

| | | Event Representations | | | | | | |
|---|---|---|---|---|---|---|---|---|
| | Configuration | Histogram [11] | MDES [12] | Voxel Grid [16] | TORE [3] | T. Surface [9] | ERGO-12 [17] | Tencode [8] |
| | Baseline | 13.33 | 12.07 | 12.75 | 12.23 | 11.69 | 11.58 | 11.15 |
| (A) | VSH | 12.29 | 10.33 | 11.71 | 10.49 | 10.20 | 10.07 | 9.84 |
| (B) | (A)+Occlusion handling [4] | 12.28 | 10.33 | 11.72 | 10.50 | 10.19 | 10.06 | 9.83 |
| (C) | (B)+Splatting [4] | 12.00 | 10.33 | 11.70 | 10.51 | 10.14 | 10.04 | 9.82 |

| | | Event Representations | | | | | | |
|---|---|---|---|---|---|---|---|---|
| | Configuration | Histogram [11] | MDES [12] | Voxel Grid [16] | TORE [3] | T. Surface [9] | ERGO-12 [17] | Tencode [8] |
| | Baseline | 13.33 | 12.07 | 12.75 | 12.23 | 11.69 | 11.58 | 11.15 |
| (A) | BTH | 10.99 | 10.53 | 10.52 | 10.34 | 10.13 | 10.09 | 9.78 |
| (B) | (A)+Occ. disc. [4] | 10.99 | 10.52 | 10.53 | 10.33 | 10.12 | 10.08 | 9.77 |
| (C) | (B)+Splatting | 11.00 | 10.54 | 10.53 | 10.34 | 10.13 | 10.08 | 9.77 |

## 4  Ablation Study – Additional Experiments

We complement the ablation studies already shown in the main paper. Specifically, Tab. I shows the results achieved by VSH and BTH on the search split, respectively on top and bottom.

Starting from VSH, we show how handling occlusions according to [4] (B) allows for slightly improving the results, with further improvements achieved by applying sub-pixel splitting [4] (C). When it comes to BTH, we observed empirically that discharging occlusions yields the best results (B). We also implemented a revised version of pixel splatting applied to the event streams (C), yet without noticeable improvements.

29

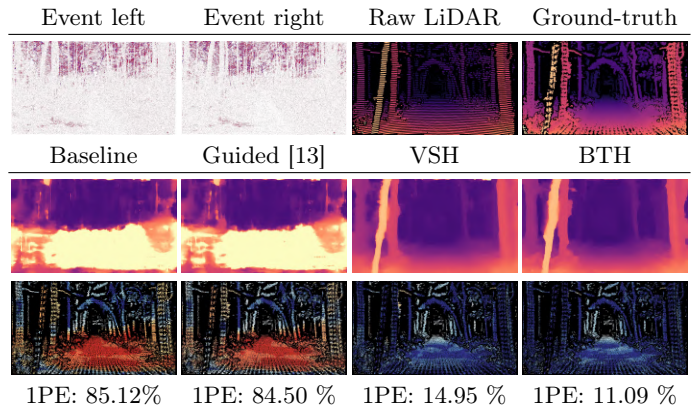

1PE: 85.12%   1PE: 84.50 %   1PE: 14.95 %   1PE: 11.09 %

**Fig. XI: Qualitative results – baseline stereo backbone.** Results on *car_forest_tree_tunnel*, with MDES [12] representation.

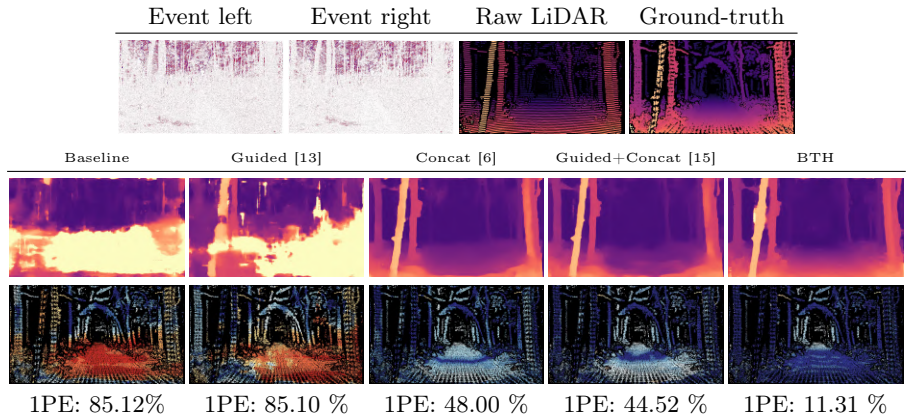

1PE: 85.12%   1PE: 85.10 %   1PE: 48.00 %   1PE: 44.52 %   1PE: 11.31 %

**Fig. XII: Qualitative results – re-trained stereo backbone.** Results on *car_forest_tree_tunnel*, with MDES [12] representation.

## 5 Qualitative Results

In conclusion, we present some qualitative results to support the efficacy of our proposal. We display eight (*i.e.*, Figs. XI to XVIII) different figures from the M3ED [5] dataset, using raw LiDAR measurements as guidance for all fusion frameworks.

Figs. XI and XII show an example from the *car_forest_tree_tunnel* sequence, respectively spotlighting fusion strategies applied without re-retraining the stereo backbone (the former) or when the network is trained from scratch to perform fusion (the latter), processing MDES [12] representations in both cases. In the former case, Guided [13] is nearly ineffective, whereas both VSH and BTH largely improve the results. In the latter case, Concat [6] and Guided+Concat [15] can reduce the error by about 40%, yet far behind the improvement yielded by BTH (more than 70% error rate reduction).

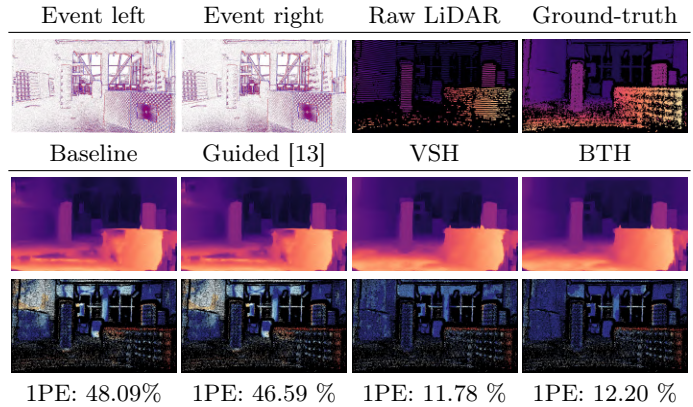

Fig. XIII: **Qualitative results – baseline stereo backbone.** Results on *spot_indoor_obstacles*, with Voxel Grid [16] representation.

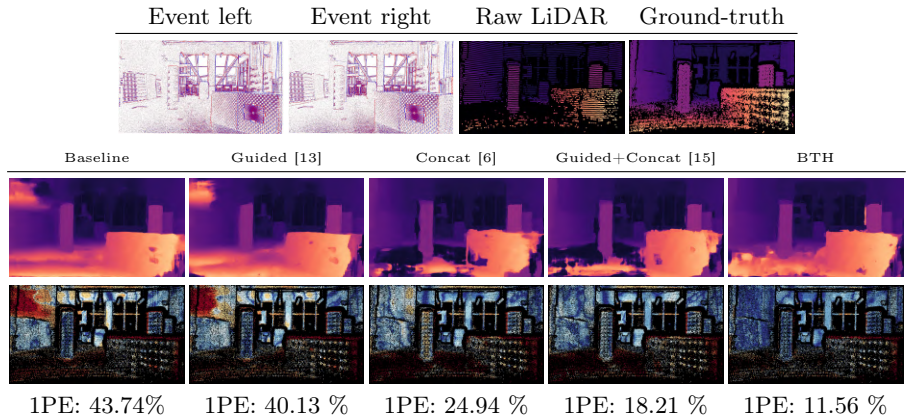

Fig. XIV: **Qualitative results – re-trained stereo backbone.** Results on *spot_indoor_obstacles*, with ERGO-12 [17] representation.

Figs. XIII and XIV show an example from an indoor sequence – *i.e.*, *spot_indoor_obstacles* – again when not re-training or training the stereo backbone from scratch. In the former case, we report results by processing Voxel Grid representations [16], which confirm the trend observed in the previous example. Indeed, our proposal confirms again the best solution for exploiting raw LiDAR measurements and improve the accuracy of event-based stereo networks.

Figs. XV and XVI showcase another example from outdoor, with a person being framed by the event cameras – *i.e.*, *spot_outdoor_day_skatepark_1*. In this case, we report results obtained by processing TORE [3] representations when re-training the stereo backbones, confirming again how our proposal consistently yields the largest drop of the error rate independently of the chosen representation.

To conclude, we evaluate the effectiveness of different fusion strategies against partially filled raw LiDAR depth maps both in indoor (*i.e.*, *falcon_indoor_-*

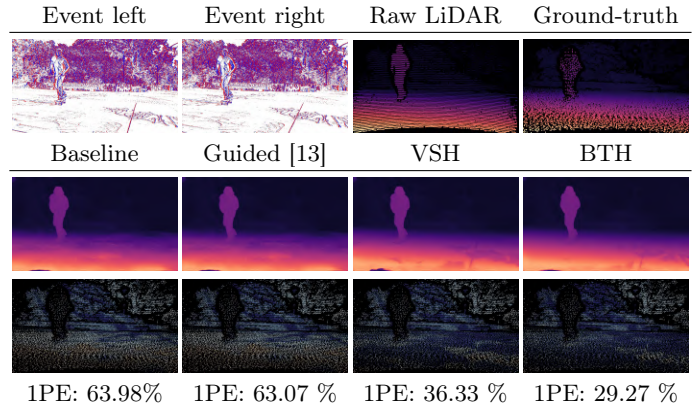

**Fig. XV: Qualitative results – baseline stereo backbone.** Results on *spot_outdoor_day_skatepark_1*, with MDES [12] representation.

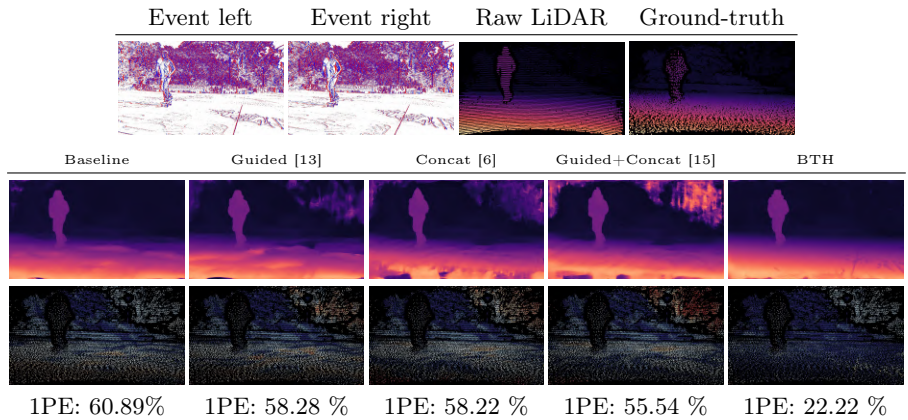

**Fig. XVI: Qualitative results – re-trained stereo backbone.** Results on *spot_outdoor_day_skatepark_1*, with TORE [3] representation.

*flight_1*) and outdoor (*i.e.*, *falcon_outdoor_day_penno_parking_1*) scenarios. The former case (Fig. XVII) highlights the behaviour of not-retrained stereo backbones at recovering fine details such as the tip of the cone and the upper part of the left-most foreground object. Compared to other methodologies, BTH manages to preserve more thin details. Furthermore, the latter case (Fig. XVIII) stress the performance of retrained stereo backbones in case of large uniform areas – *i.e.*, the road. Guided [13] is almost ineffective, while using both Concat [6] and Guided+Concat [15] leads to 20% error reduction. However, when dealing with large homogeneous regions where LiDAR coverage is limited, BTH clearly dominates alternative methods, achieving a remarkable 80% reduction in error.

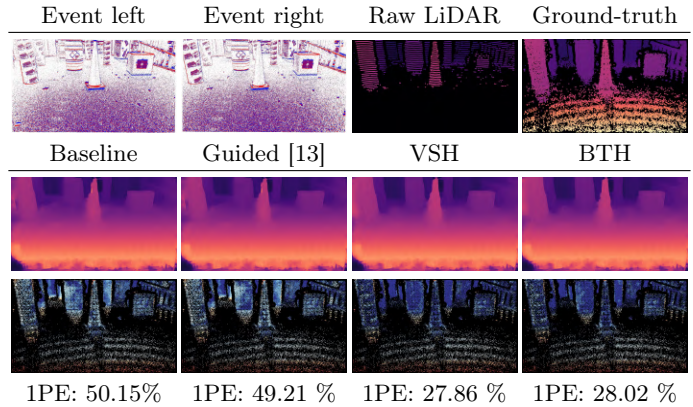

**Fig. XVII: Qualitative results – baseline stereo backbone.** Results on *falcon_indoor_flight_1*, with Voxel Grid [16] representation.

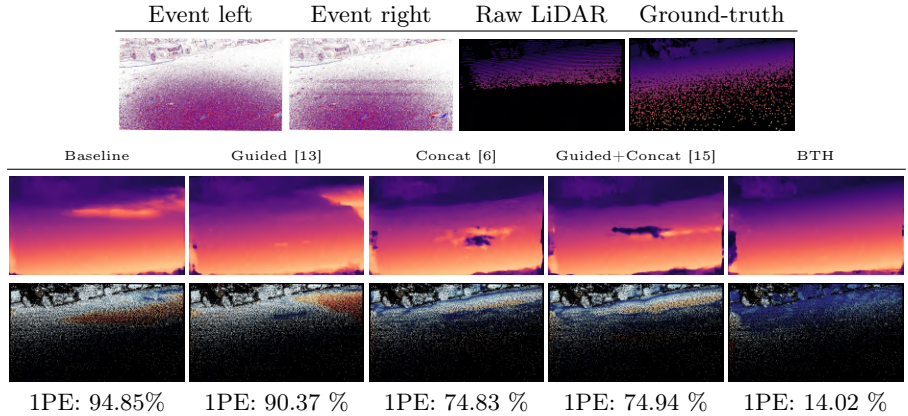

**Fig. XVIII: Qualitative results – re-trained stereo backbone.** Results on *falcon_outdoor_day_penno_parking_1*, with MDES [12] representation.


}

\end{document}